\documentclass{article}

\usepackage[preprint]{neurips_2026}

\usepackage[utf8]{inputenc} % allow utf-8 input
\usepackage[T1]{fontenc}    % use 8-bit T1 fonts
\usepackage{hyperref}       % hyperlinks
\usepackage{url}            % simple URL typesetting
\usepackage{booktabs}       % professional-quality tables
\usepackage{amsfonts}       % blackboard math symbols
\usepackage{nicefrac}       % compact symbols for 1/2, etc.
\usepackage{microtype}      % microtypography
\usepackage{xcolor}         % colors
\usepackage{algorithm}
\usepackage{algpseudocode}
\usepackage{caption}
\usepackage{wrapfig}
\usepackage{enumitem}
\usepackage{subcaption}
\usepackage[table]{xcolor}
\usepackage{graphicx}
\usepackage[normalem]{ulem}
\usepackage{float}
\usepackage{standalone}
\usepackage{chngcntr}
\usepackage{makecell}
\usepackage{amsmath}
\usepackage{amsfonts}

\usepackage{tikz}
\usepackage{xcolor}
\usepackage{graphicx}
\usepackage{caption}
\usetikzlibrary{positioning,arrows.meta,fit}
\usetikzlibrary{backgrounds}
\usetikzlibrary{positioning,calc,fit,arrows.meta}

\definecolor{paperBlue}{RGB}{47,82,143}
\definecolor{paperLightBlue}{RGB}{232,239,249}
\definecolor{paperOrange}{RGB}{197,111,44}
\definecolor{paperLightOrange}{RGB}{252,238,221}
\definecolor{paperGray}{RGB}{246,246,246}
\definecolor{paperDarkGray}{RGB}{70,70,70}
\newcommand{\bsig}[1]{\cellcolor{green!15}\textbf{#1}}

\DeclareMathOperator{\LN}{LN}
\DeclareMathOperator{\softmax}{softmax}

\AtBeginDocument{%
  }

\usepackage[most]{tcolorbox}
\newtcolorbox{rodbox}[1]{%
  colback=#1!50,
  colframe=#1!80!black,
  fontupper=\normalsize,
  left=1mm, right=1mm, top=0.5mm, bottom=0.5mm,
  boxrule=0.5pt,
  arc=1pt,
  width=\linewidth,
  boxsep=1mm,
  before skip=6pt, after skip=6pt,
  enhanced,
}

\title{Inference-Time Decision Calibration\\for Temporal Classification}

\author{%
  \textbf{Arthur Chagas}\thanks{Corresponding Author},
  \textbf{Arthur Buzelin},
  \textbf{Yan Aquino},
  \textbf{Pedro Augusto Torres Bento}, \\
  \textbf{Gisele L.~Pappa},
  \textbf{Wagner Meira Jr.},
  \textbf{Cristiano Arbex Valle} \\
  Department of Computer Science (DCC), Universidade Federal de Minas Gerais (UFMG) \\
  \texttt{arthurchagas@dcc.ufmg.br}
}

\begin{document}

\maketitle

\begin{abstract}
Temporal classification errors are often treated as representation failures, but they can also arise from how available evidence is converted into decisions. This paper proposes a representation--calibration decomposition for temporal classification. We keep a trained native classifier frozen and separate two inference-time interventions: a conservative residual multi-scale branch that adds auxiliary logits to the native prediction, and a post-hoc branch-aware calibrator that recombines native and residual evidence at decision time. This design distinguishes missing temporal evidence from underused decision-level evidence without retraining the backbone. Across FI-2010, PTB-XL, UCI-HAR, MHEALTH, and HARTH, we find that gains are strongly regime-dependent. Residual multi-scale evidence is most useful in noisy or representation-limited settings, especially short-horizon FI-2010 and weaker recurrent backbones, while branch-aware calibration helps when native and auxiliary logits contain complementary evidence not fully exploited by the raw decision rule. Near-saturated settings show limited gains from either intervention. These results suggest that temporal classification should be understood not only as representation learning, but also as the problem of trusting, combining, and calibrating evidence from multiple views.
\end{abstract}

\section{Introduction}

Temporal classification is usually treated as a representation problem: if a model fails, the natural response is to build a stronger temporal encoder. This view has driven increasingly expressive recurrent, convolutional, residual, Transformer-based, and multi-scale classifiers for sequential data \citep{bagnall,ruizetal,hivecote,Tan2022}. But a trained classifier does not make decisions from representations directly. It makes them from logits, thresholds, branch weights, and calibration choices. When the target metric is macro-F1 rather than the training loss, this final score-to-decision step can change performance without changing the temporal representation at all \citep{Vaicenavicius2019EvaluatingMC,minderer,NEURIPS2019_f8c0c968,yenan,parambath,dembczynski}.

This paper asks the following question: when a temporal classifier still makes errors, is it missing temporal evidence, or is it misusing evidence already present in its logits? Standard approaches make this hard to answer, because replacing the backbone or jointly training a larger model changes representation learning and decision conversion at the same time. We instead freeze a trained temporal classifier and separate two inference-time interventions: a conservative residual multi-scale correction to its logits, and a post-hoc calibrator fitted over frozen native and auxiliary evidence. Unlike test-time adaptation methods that update model parameters on unlabeled test inputs \citep{yusun,wang2021tent,zhang2022memo}, our calibration stage changes only the decision rule.

This gives a direct diagnostic decomposition. If the raw residual branch improves performance, the native model lacked useful temporal evidence. If raw residual gains are small but calibration improves the final prediction, useful evidence was present but underused at the decision layer. If neither helps, the dataset or backbone is close to saturation. The goal is therefore not to propose a universal residual module that wins everywhere, but to identify which bottleneck dominates in each temporal classification regime.

Empirically, the decomposition reveals that residual and calibration gains are strongly regime-dependent. Residual multi-scale evidence is most valuable when the native predictor is representation-limited or exposed to noisy temporal structure. Calibration is most useful when native and auxiliary logits contain complementary evidence that is not optimally combined by the raw decision rule. In contrast, strong backbones and near-saturated settings often leave little room for either correction. These findings suggest that temporal classification errors should not be attributed only to weak temporal encoders: in many cases, the bottleneck lies in how available evidence is combined and converted into final decisions.

Our contributions are:
\begin{enumerate}
    \item We propose a \emph{representation--calibration decomposition} for temporal classification, separating residual error into missing temporal evidence, underused decision-level evidence, and near-saturation regimes.
    
    \item We introduce a \emph{conservative residual multi-scale adapter} that keeps the native temporal predictor frozen and adds auxiliary logits as an additive correction, enabling cleaner analysis of complementary temporal evidence without backbone retraining.
    
    \item We study \emph{branch-aware post-hoc decision calibration} over frozen native and auxiliary logits, allowing us to test whether residual evidence is useful even when raw residual gains are small.
    
    \item We provide an \emph{empirical regime analysis} across five temporal classification benchmarks and multiple backbone families.
\end{enumerate}

\section{Related Work}
\label{sec:related_work}

\paragraph{Temporal representation learning.}
Temporal classification is commonly treated as a representation-learning problem: better sequence encoders should extract more discriminative temporal evidence from raw sequential observations. Strong time-series classifiers include convolutional architectures such as InceptionTime~\citep{inceptiontime} and ROCKET~\citep{dempster2019rocket}, as well as Transformer-based models adapted from general sequence modeling~\citep{NIPS2017_3f5ee243} to multivariate time-series representation learning~\citep{temporal}. Recent time-series architectures further emphasize decomposition, patching, and long-context aggregation~\citep{NEURIPS2021_bcc0d400,patch}. These works mainly ask how to improve the temporal encoder. Our work asks a complementary question: once a strong temporal predictor is already available, are remaining errors caused only by missing representations, or also by how available evidence is converted into final decisions?

\paragraph{Multi-scale and residual temporal evidence.}
Many temporal domains contain information at multiple resolutions: local windows may capture short-lived events, while coarser views may provide more stable context. Temporal Fusion Transformers combine local processing, attention, and gating to integrate temporal evidence across horizons~\citep{LIM20211748}. Residual learning provides a general mechanism for improving a predictor by learning corrections around an existing solution~\citep{resnet}. Our method combines these ideas, but uses them conservatively: the native temporal predictor remains explicit, and the multi-scale branch contributes only additive residual logits. This design differs from replacing the backbone with a larger architecture, because it lets us distinguish gains due to additional temporal evidence from gains due to better decision conversion.

\paragraph{Post-hoc calibration and inference-time decisions.}
Post-hoc calibration adjusts classifier outputs after training without relearning the representation. Temperature scaling and related neural calibration methods have shown that modern neural networks can be accurate but poorly calibrated~\citep{pmlr-v70-guo17a}, while Dirichlet calibration and verified calibration extend this line to richer multiclass and reliability-aware settings~\citep{NEURIPS2019_8ca01ea9,NEURIPS2019_f8c0c968}. Calibration is especially relevant when the decision metric, such as macro-F1, is not perfectly aligned with the training loss; thresholding analyses for F-measure optimization show that final decision rules can matter independently of score quality~\citep{10.1007/978-3-662-44851-9_15}. Our approach is also related to test-time adaptation, which modifies models at deployment time~\citep{wang2021tent}, but differs in a key respect: we freeze the temporal representation and fit a lightweight branch-aware decision layer over native and residual logits. Thus, post-calibration gains are interpreted as improved decision conversion rather than new representation learning.

\section{Preliminaries and Definitions}
\label{sec:preliminaries}

We study supervised temporal classification. Each input is a multivariate sequence
\(
x=(x_1,\dots,x_T)\in\mathbb{R}^{T\times D},
\)
where \(T\) is the temporal length and \(D\) is the number of input channels. For single-label datasets, the label is
\(
y\in\{1,\dots,K\},
\)
where \(K\) is the number of classes. A classifier outputs logits
\(
\ell(x)\in\mathbb{R}^K,
\)
which induce the predictive distribution
\[
p(x)=\softmax(\ell(x))\in\Delta^{K-1}.
\]
The predicted class is
\(
\hat y(x)=\arg\max_{k\in\{1,\dots,K\}}\ell_k(x).
\)

For multi-label datasets, such as PTB-XL in our experiments, labels are vectors
\(
y\in\{0,1\}^K.
\)
The classifier again outputs logits
\(
\ell(x)\in\mathbb{R}^K,
\)
but each class is converted independently into a sigmoid probability:
\(
p_k(x)=\sigma(\ell_k(x)),
\space k=1,\dots,K.
\)
Binary decisions are obtained as
\(
\hat y_k(x)
=
\mathbf{1}\{p_k(x)\ge \theta_k\},
\space k=1,\dots,K,
\)
where \(\theta_k=0.5\) unless otherwise specified. Single-label models are trained with cross-entropy, while multi-label models are trained with binary cross-entropy with logits.

A temporal encoder maps the input sequence to latent states
\(
H(x)=(h_1(x),\dots,h_T(x))\in\mathbb{R}^{T\times m},
\)
where \(m\) is the hidden dimension. A readout operator aggregates the hidden sequence into a fixed-dimensional representation
\(
z(x)\in\mathbb{R}^{m}.
\)

When multiple temporal resolutions are considered, we denote by
\(
\mathcal{S}=\{s_1,\dots,s_M\},
\space s_1=1,
\)
the set of scale factors. The scale \(s_1=1\) corresponds to the original temporal resolution, while larger scale factors correspond to progressively coarser temporal views. For each scale \(s\in\mathcal{S}\), we write
\(
x^{(s)}=P_s(x)\in\mathbb{R}^{T_s\times D}
\)
for the projected sequence, with \(T_s\le T\).

Throughout the paper, \(\ell_{\mathrm{nat}}(x)\) denotes the native logits, \(\ell_{\mathrm{aux}}(x)\) denotes the auxiliary residual logits, \(\ell_{\mathrm{raw}}(x)\) denotes the uncalibrated combined logits, and \(\tilde{\ell}(x)\) denotes the calibrated logits. The branch-wise evidence vector used by branch-aware calibrators is denoted
\(
v(x)
=
[\ell_{\mathrm{nat}}(x);\ell_{\mathrm{aux}}(x)]
\in\mathbb{R}^{2K}.
\)
Scale-trust weights are denoted by \(\alpha_s(x)\), with
\(
\alpha_s(x)\ge 0,
\space
\sum_{s\in\mathcal{S}}\alpha_s(x)=1.
\)

We reserve \(\gamma_{\mathrm{res}}\in[0,1]\) for the scalar residual weight used when combining native and auxiliary logits, and \(\Gamma\in\mathbb{R}^K\) for class-wise residual gains used by branch-aware calibration. The symbol \(\odot\) denotes element-wise multiplication, \(\|\cdot\|_2\) denotes the Euclidean norm, \(\sigma(\cdot)\) denotes the logistic sigmoid, and \(\LN(\cdot)\) denotes layer normalization.

\section{Method}
\label{sec:method}

\paragraph{Problem setup.}
Let
\(
\mathcal{D}=\{(x_i,y_i)\}_{i=1}^N
\)
be a labeled temporal classification dataset, where each input sequence is
\(
x_i\in\mathbb{R}^{T\times D}.
\)
Our starting point is a native temporal classifier. The central hypothesis of this paper is that, once such a backbone is already competent, the remaining error need not be purely representational. Part of it may instead arise from how native evidence and auxiliary temporal evidence are combined into the final decision. We therefore decompose the predictor into two components:
\begin{enumerate}
    \item a \emph{native evidence branch}, which preserves the original backbone prediction;
    \item a \emph{residual multi-scale branch}, which provides an additive auxiliary correction;
\end{enumerate}
followed by an optional \emph{post-hoc decision calibration} stage that operates after representation learning is frozen.

\begin{figure}[t]
\centering

\begin{minipage}[t]{0.58\textwidth}
\vspace{0pt}
\centering
\begin{tikzpicture}[
    font=\large,
    box/.style={
        rectangle,
        rounded corners=3pt,
        thick,
        draw=paperDarkGray,
        fill=paperGray,
        align=center,
        inner sep=5.2pt,
        minimum height=2.55em,
        text width=14.4em
    },
    inputbox/.style={
        box,
        fill=paperLightBlue,
        draw=paperBlue
    },
    nativebox/.style={
        box,
        fill=paperLightBlue,
        draw=paperBlue
    },
    residualbox/.style={
        box,
        fill=paperLightOrange,
        draw=paperOrange
    },
    trustbox/.style={
        residualbox,
        text width=16.2em
    },
    fusionbox/.style={
        box,
        fill=paperGray,
        draw=paperDarkGray,
        text width=13.6em
    },
    decisionbox/.style={
        box,
        fill=paperLightBlue,
        draw=paperBlue,
        text width=16.2em
    },
    predictionbox/.style={
        box,
        fill=paperLightBlue,
        draw=paperBlue,
        text width=13.8em
    },
    arrow/.style={
        -{Latex[length=2.2mm]},
        thick,
        draw=paperDarkGray
    },
    sidearrow/.style={
        -{Latex[length=2.2mm]},
        thick,
        draw=paperDarkGray,
        rounded corners=6pt
    },
    tinylabel/.style={
        font=\normalsize,
        fill=white,
        inner sep=1.2pt,
        text=paperDarkGray
    },
    arrowlabel/.style={
        font=\normalsize,
        text=paperDarkGray,
        fill=white,
        inner sep=1pt,
        rotate=90
    }
]

% ------------------------------------------------------------
% Left column: native branch and final decision path
% ------------------------------------------------------------

\node[inputbox] (input) at (0,5.00) {
    \textbf{Temporal input}\\[-0.15em]
    $X \in \mathbb{R}^{T \times C}$
};

\node[nativebox] (native) at (0,3.30) {
    \textbf{Native Evidence}\\[-0.15em]
    $z^{(0)} = f_{\theta}(X)$\\[-0.1em]
    $\ell_{\mathrm{nat}} = W_0 z^{(0)} + b_0$
};

\node[fusionbox] (fusion) at (0,1.35) {
    \textbf{Residual fusion}\\[-0.15em]
    $\ell_{\mathrm{raw}} = \ell_{\mathrm{nat}} + \delta(X)$
};

\node[decisionbox] (calibration) at (0,-1.30) {
    \textbf{Decision layer}\\[-0.35em]
    {\normalsize
    \[
    \ell_{\mathrm{out}} =
    \begin{cases}
    \ell_{\mathrm{raw}}, & \text{raw},\\
    \mathcal C_{\varphi}([\ell_{\mathrm{nat}};\delta(X)]), & \text{calibrated}.
    \end{cases}
    \]
    }
    \vspace{-1.15em}
};

\node[predictionbox] (decision) at (0,-3.75) {
    \textbf{Prediction}\\[-0.15em]
    $\widehat y =
    \operatorname*{arg\,max}_{k}
    \ell_{\mathrm{out},k}$
};

% ------------------------------------------------------------
% Right column: residual inference
% ------------------------------------------------------------

\node[residualbox] (views) at (7.35,2.90) {
    \textbf{Multi-Scale Evidence}\\[-0.15em]
    $X_s = P_s(X)$\\[-0.1em]
    $z_s = \operatorname{Pool}(E_s(X_s))$\\[-0.1em]
    $q_s = W_s z_s + b_s$
};

\node[trustbox] (trust) at (7.35,0.25) {
    \textbf{Scale trust}\\[-0.35em]
    {\normalsize
    \[
    \alpha_s =
    \begin{cases}
    |\mathcal S|^{-1}, & \text{MS},\\
    \operatorname{softmax}_s(g_{\eta}(z_s)), & \text{ZOnly},\\
    \operatorname{softmax}_s(g_{\eta}([z_s;r_s])), & \text{RCG}.
    \end{cases}
    \]
    }
    \vspace{-1.15em}
};

\node[residualbox] (aux) at (7.35,-2.20) {
    \textbf{Auxiliary correction}\\[-0.15em]
    $\ell_{\mathrm{aux}} \equiv \delta(X)$\\[-0.1em]
    $\delta(X) = \sum_{s \in \mathcal S}\alpha_s q_s$
};

% ------------------------------------------------------------
% Arrows
% ------------------------------------------------------------

\draw[arrow] (input) -- (native);
\draw[arrow] (native) -- (fusion);

\draw[sidearrow] (native.east) -- ++(1.00,0) |- (views.west)
    node[pos=0.18,above=0.05cm,arrowlabel] {$X$ reused};

\draw[arrow] (views) -- (trust);
\draw[arrow] (trust) -- (aux);

\draw[sidearrow] (aux.west) -- ++(-1.20,0) |- (fusion.east)
    node[pos=0.22,below=0.05cm,arrowlabel] {$\delta(X)$};

\draw[arrow] (fusion) -- (calibration);
\draw[arrow] (calibration) -- (decision);

% ------------------------------------------------------------
% Enclosing regions
% ------------------------------------------------------------

\begin{scope}[on background layer]
\node[
    draw=paperBlue,
    dashed,
    thick,
    rounded corners=4pt,
    inner xsep=9pt,
    inner ysep=10pt,
    fit=(input)(native),
    label={[tinylabel, text=paperBlue]above:\textbf{native evidence}}
] {};

\node[
    draw=paperOrange,
    dashed,
    thick,
    rounded corners=4pt,
    inner xsep=9pt,
    inner ysep=10pt,
    fit=(views)(trust)(aux),
    label={[tinylabel, text=paperOrange]above:\textbf{residual inference}}
] {};

\node[
    draw=paperBlue,
    dashed,
    thick,
    rounded corners=4pt,
    inner xsep=9pt,
    inner ysep=10pt,
    fit=(calibration)(decision),
    label={[tinylabel, text=paperBlue]below:\textbf{decision conversion}}
] {};
\end{scope}

\end{tikzpicture}
\end{minipage}
\hfill
\begin{minipage}[t]{0.40\textwidth}
\vspace{0pt}
\begingroup
\setlength{\columnwidth}{\linewidth}

\begin{algorithm}[H]
\small
\caption{Inference-Time Decision Calibration}
\label{alg:branch_calibration}
\begin{algorithmic}[1]
\Require trained residual model, validation set \(\mathcal{D}_{\mathrm{val}}\), calibrator \(\mathcal{C}_{\phi}\)
\Ensure calibrated prediction \(\hat y\)

\State Freeze all representation parameters
\State Compute \(v_i=[\ell_{\mathrm{nat}}(x_i);\ell_{\mathrm{aux}}(x_i)]\) for all \((x_i,y_i)\in\mathcal{D}_{\mathrm{val}}\)
\State Fit \(\phi^\star\) using \(\{(v_i,y_i)\}_{i=1}^{|\mathcal{D}_{\mathrm{val}}|}\)
\State For test input \(x\), compute \(v(x)=[\ell_{\mathrm{nat}}(x);\ell_{\mathrm{aux}}(x)]\)
\State \(\tilde{\ell}(x)=\mathcal{C}_{\phi^\star}(v(x))\)
\State \Return \(\hat y=\arg\max_k \tilde{\ell}_k(x)\) for single-label tasks, or thresholded sigmoid decisions for multi-label tasks

\end{algorithmic}
\end{algorithm}

\endgroup
\end{minipage}

\caption{Overview of the representation--calibration decomposition.}
\label{fig:method_diagram}
\end{figure}

\subsection{Native--Residual Decomposition}

Let \(f_{\psi}\) denote the native backbone, with parameters \(\psi\). Given an input sequence \(x\), the backbone produces native logits
\(
\ell_{\mathrm{nat}}(x)
=
f_{\psi}(x)
\in\mathbb{R}^{K}.
\)
When the backbone exposes an intermediate representation, we write this representation as
\(
z_{\mathrm{nat}}(x)=h_{\psi}(x),
\space
\ell_{\mathrm{nat}}(x)
=
W_{\mathrm{nat}}z_{\mathrm{nat}}(x)+b_{\mathrm{nat}}.
\)
This branch is the reference predictor and remains explicit throughout the method.

We then construct a residual multi-scale branch that extracts auxiliary evidence from several temporal resolutions. Let
\(
\mathcal{S}=\{s_1,\dots,s_M\},
\space s_1=1,
\)
be the ordered set of temporal scales. For each scale \(s\in\mathcal{S}\), we apply a scale operator \(P_s\), implemented as temporal downsampling or average pooling, to obtain
\(
x^{(s)}
=
P_s(x)
\in\mathbb{R}^{T_s\times D}.
\)
Each scale is encoded by a scale-specific encoder \(E_s\), yielding hidden states
\(
H^{(s)}
=
E_s(x^{(s)})
\in\mathbb{R}^{T_s\times d}.
\)
We summarize each scale into a fixed-size vector
\[
z^{(s)}
=
\operatorname{Pool}\!\left(H^{(s)}\right)
\in\mathbb{R}^{d}.
\]
In implementations with cross-scale representation attention, the scale vectors are refined jointly:
\[
(\tilde z^{(s_1)},\dots,\tilde z^{(s_M)})
=
A_{\omega}\!\left(z^{(s_1)},\dots,z^{(s_M)}\right).
\]
In the generic cross-backbone experiments, \(A_{\omega}\) is the identity map, so \(\tilde z^{(s)}=z^{(s)}\).

The residual branch does not replace the native predictor. Instead, it aggregates scale representations into an auxiliary representation
\(
z_{\mathrm{aux}}(x)
=
\sum_{s\in\mathcal{S}}
\alpha_s(x)\tilde z^{(s)}(x),
\)
where the scale-trust weights satisfy
\(
\alpha_s(x)\ge 0,
\qquad
\sum_{s\in\mathcal{S}}\alpha_s(x)=1.
\)
The auxiliary representation is then mapped to auxiliary logits:
\(
\ell_{\mathrm{aux}}(x)
=
W_{\mathrm{aux}}z_{\mathrm{aux}}(x)+b_{\mathrm{aux}}.
\)
The raw combined logits are
\(
\ell_{\mathrm{raw}}(x)
=
\ell_{\mathrm{nat}}(x)
+
\gamma_{\mathrm{res}}\ell_{\mathrm{aux}}(x),
\space
\gamma_{\mathrm{res}}=\sigma(\rho).
\)
This additive structure is important: it forces auxiliary evidence to act as a correction to the native decision, rather than allowing the auxiliary branch to overwrite the entire model.

\subsection{Trust Allocation Across Temporal Scales}

The trust weights \(\alpha_s(x)\) determine how much each temporal scale contributes to the auxiliary correction. We study three variants.

\paragraph{Residual-MS.}
The simplest variant uses uniform fusion:
\(
\alpha_s(x)
=
\frac{1}{M},
\space
\forall s\in\mathcal{S}.
\)
This tests whether multi-scale residual evidence is useful even without learned scale selection.

\paragraph{Residual-ZOnly.}
The second variant learns trust from scale representations alone. Given scale representations
\(
\{\tilde z^{(s)}(x)\}_{s\in\mathcal{S}},
\)
a small gating network produces scale scores
\(
a_s(x)
=
g_{\theta}\!\left(\tilde z^{(s)}(x)\right).
\)
The adaptive weights are
\[
\alpha^{\mathrm{ad}}_s(x)
=
\frac{\exp(a_s(x)/\tau)}
{\sum_{r\in\mathcal{S}}\exp(a_r(x)/\tau)}.
\]
When uniform interpolation is enabled, the final trust weights are
\(
\alpha_s(x)
=
(1-\mu)\frac{1}{M}
+
\mu\,\alpha^{\mathrm{ad}}_s(x),
\space
\mu=\sigma(\eta).
\)
When this interpolation is not used, this reduces to
\(
\alpha_s(x)=\alpha^{\mathrm{ad}}_s(x).
\)
This variant asks whether scale selection can be learned from representations alone.

\paragraph{Residual-RCG.}
The third variant augments the scale representations with reliability descriptors. In the latent reliability version, for each scale \(s\), we compute
\(
r^{(s)}(x)
=
\big[
e^{(s)}(x),\;
u^{(s)}(x),\;
c^{(s)}(x),\;
\nu^{(s)}(x)
\big],
\)
where the components measure local instability, recent instability, cross-scale consistency, and representation energy, respectively. With
\(
H^{(s)}\in\mathbb{R}^{T_s\times d},
\)
these descriptors are
\begin{align}
e^{(s)}(x)
&=
\frac{1}{(T_s-1)d}
\sum_{t=2}^{T_s}
\sum_{j=1}^{d}
\left(
H^{(s)}_{t,j}
-
H^{(s)}_{t-1,j}
\right)^2,
\\
u^{(s)}(x)
&=
\frac{1}{w_s d}
\sum_{t=T_s-w_s+1}^{T_s}
\sum_{j=1}^{d}
\left(
H^{(s)}_{t,j}
-
\bar H^{(s)}_{\mathrm{rec},j}
\right)^2,
\\
\nu^{(s)}(x)
&=
\frac{1}{d}
\sum_{j=1}^{d}
\left(\tilde z^{(s)}_j(x)\right)^2.
\end{align}
Here \(w_s\) is the recent-window length and
\(
\bar H^{(s)}_{\mathrm{rec}}
=
\frac{1}{w_s}
\sum_{t=T_s-w_s+1}^{T_s}
H^{(s)}_t
\)
is the average hidden state over the last \(w_s\) positions. Cross-scale consistency is computed from neighboring scale summaries:
\[
c^{(s)}(x)
=
-\frac{1}{|\mathcal{N}(s)|}
\sum_{r\in\mathcal{N}(s)}
\frac{1}{d}
\sum_{j=1}^{d}
\left(
\tilde z^{(s)}_j(x)
-
\tilde z^{(r)}_j(x)
\right)^2,
\]
where \(\mathcal{N}(s)\) contains the adjacent scales of \(s\). Because \(c^{(s)}(x)\) is the negative average squared distance to neighboring scales, larger values indicate stronger agreement with neighboring resolutions. The descriptors are standardized before entering the gate.

The scale scores are then computed from both representations and reliability descriptors:
\[
a_s(x)
=
g_{\theta}\!\left([\tilde z^{(s)}(x);r^{(s)}(x)]\right),
\]
followed by the same softmax, and optional uniform interpolation, described above. In the generic cross-backbone experiments, RCG uses an analogous descriptor vector computed directly from each input-scale view, including input energy, amplitude, standard deviation, and first-difference magnitude. This variant tests whether explicit reliability signals improve trust allocation beyond scale embeddings alone.

\subsection{Training Objective}

The native classifier is trained first and then kept fixed. The residual branch is trained afterward on top of the frozen native logits. Gradients update only the residual encoders, scale-trust module, auxiliary head, and residual weight \(\gamma_{\mathrm{res}}\).

For single-label datasets, the residual stage is optimized using cross-entropy on the combined logits:
\[
\mathcal{L}_{\mathrm{raw}}
=
-\frac{1}{N}
\sum_{i=1}^{N}
\log
\frac{
\exp(\ell_{\mathrm{raw},y_i}(x_i))
}{
\sum_{k=1}^{K}
\exp(\ell_{\mathrm{raw},k}(x_i))
}.
\]
For multi-label datasets, the same stage uses binary cross-entropy with logits applied to \(\ell_{\mathrm{raw}}(x_i)\).

When auxiliary residual regularization is enabled, we add an auxiliary classification loss on \(\ell_{\mathrm{aux}}\), a scale-consistency loss, and an entropy regularization term:
\[
\mathcal{L}_{\mathrm{res}}
=
\mathcal{L}_{\mathrm{raw}}
+
\lambda_{\mathrm{aux}}\mathcal{L}_{\mathrm{aux}}
+
\lambda_{\mathrm{cons}}\mathcal{L}_{\mathrm{cons}}
+
\lambda_{\mathrm{ent}}\mathcal{L}_{\mathrm{ent}}.
\]
Here \(\mathcal{L}_{\mathrm{aux}}\) is the same supervised loss applied to \(\ell_{\mathrm{aux}}\). The consistency and entropy terms are
\[
\mathcal{L}_{\mathrm{cons}}
=
\frac{1}{N}
\sum_{i=1}^{N}
\frac{1}{M-1}
\sum_{m=1}^{M-1}
\left\|
\tilde z^{(s_m)}(x_i)
-
\tilde z^{(s_{m+1})}(x_i)
\right\|_2^2,
\]
and
\[
\mathcal{L}_{\mathrm{ent}}
=
\frac{1}{N}
\sum_{i=1}^{N}
\sum_{s\in\mathcal{S}}
\alpha_s(x_i)\log\alpha_s(x_i).
\]
This makes the residual branch a conservative correction to a fixed predictor, rather than a joint retraining of the native and auxiliary branches.

\subsection{Post-hoc Decision Calibration}

After training the raw predictor, all representation parameters are frozen. We then study whether additional gains can be obtained purely at the \emph{decision level}, without changing the learned temporal features. This is the second part of the decomposition.

For residual models, let
\(
v(x)
=
[\ell_{\mathrm{nat}}(x);\ell_{\mathrm{aux}}(x)]
\in\mathbb{R}^{2K}
\)
denote the branch-wise evidence vector, which preserves the distinction between native and auxiliary evidence. A post-hoc calibrator is a map
\(
\mathcal{C}_{\phi}:\mathbb{R}^{2K}\to\mathbb{R}^{K},
\)
with calibrated logits
\(
\tilde{\ell}(x)
=
\mathcal{C}_{\phi}(v(x)).
\)

For single-label datasets,
\(
p_{\phi}(y\mid x)
=
\softmax\!\left(\tilde{\ell}(x)\right)_y,
\space
\hat y(x)
=
\arg\max_k \tilde{\ell}_k(x).
\)
For multi-label datasets, calibrated logits are converted to independent sigmoid probabilities:
\(
p_{\phi,k}(x)
=
\sigma(\tilde{\ell}_k(x)),
\space
\hat y_k(x)
=
\mathbf{1}\{p_{\phi,k}(x)\ge \theta_k\}.
\)

This formulation is intentionally general. In experiments, we instantiate \(\mathcal{C}_{\phi}\) with both branch-agnostic and branch-aware post-hoc rules. Branch-agnostic calibrators act on \(\ell_{\mathrm{raw}}(x)\), including temperature scaling, vector scaling, matrix scaling, isotonic regression, histogram binning, and final-logit search. Branch-aware calibrators operate on the separated native and auxiliary logits. In the separated search-based form, the calibrated logits are
\[
\tilde{\ell}(x)
=
\frac{\ell_{\mathrm{nat}}(x)}{T_{\mathrm{nat}}}
+
\Gamma\odot
\left(
\frac{\ell_{\mathrm{aux}}(x)}{T_{\mathrm{aux}}}
\right)
+
b_{\mathrm{cal}},
\]
where \(T_{\mathrm{nat}}>0\) and \(T_{\mathrm{aux}}>0\) are positive temperatures, \(\Gamma\in\mathbb{R}^{K}\) is a learned class-wise residual gain vector, \(b_{\mathrm{cal}}\in\mathbb{R}^{K}\) is a calibration bias vector, and \(\odot\) denotes element-wise multiplication.

For differentiable calibration families, \(\phi\) is fit on a validation set by minimizing the corresponding negative log-likelihood. In the single-label case,
\[
\mathcal{L}_{\mathrm{cal}}(\phi)
=
-\frac{1}{|\mathcal{D}_{\mathrm{val}}|}
\sum_{(x_i,y_i)\in \mathcal{D}_{\mathrm{val}}}
\log p_{\phi}(y_i\mid x_i).
\]
For multi-label datasets, the calibration loss is binary cross-entropy with logits applied to \(\tilde{\ell}(x_i)\). For search-based calibrators, \(\phi\) is selected directly against validation macro-F1. In all cases, calibration is learned \emph{after} freezing the temporal representation, so any post-calibration improvement can be attributed to better decision conversion rather than better representation learning.

\subsection{Interpretation of the Decomposition}

The method induces a simple diagnostic interpretation:
\begin{itemize}
    \item if the raw residual branch improves over the native model, the bottleneck is at least partly \emph{representational};
    \item if raw gains are small but post-hoc calibration improves final performance, the bottleneck is at least partly \emph{decisional};
    \item if neither intervention helps, the task is likely in a near-saturation regime for the given backbone.
\end{itemize}
This interpretation is central to the paper. The goal is to make explicit the distinction between \emph{missing evidence} and \emph{underused evidence} in temporal classification.

\section{Experiments}
\label{sec:experiments}

We evaluate the proposed representation--calibration decomposition on five benchmarks with distinct temporal structure and difficulty: FI-2010\citep{fi2010} for limit-order-book prediction, PTB-XL\citep{ptbxl} for ECG classification, and UCI-HAR\citep{uci_har}, MHEALTH\citep{mhealth}, and HARTH\citep{harth} for wearable human-activity recognition. We use these datasets to diagnose where performance gains come from: additional temporal evidence, better allocation of trust across temporal scales, or improved conversion of logits into final decisions.

\paragraph{Research questions.}
Our protocol is organized around four questions:
\begin{enumerate}
    \item \textbf{RQ1: Residual temporal evidence.} Can a conservative residual multi-scale branch improve a strong native temporal classifier without replacing its original representation?
    \item \textbf{RQ2: Trust allocation.} Does learning how much to trust each temporal scale improve over uniform residual fusion, and are explicit reliability descriptors always necessary?
    \item \textbf{RQ3: Representation--decision separation.} When raw residual gains are small, do the residual branches still provide useful auxiliary evidence for post-hoc decision calibration?
    \item \textbf{RQ4: Dataset regimes.} Do the five benchmarks exhibit different regimes, such as residual-representation gains, calibration-driven gains, near-saturation, or unstable decision boundaries?
\end{enumerate}

\begin{figure}[t]
    \centering
    \includegraphics[width=0.9\textwidth]{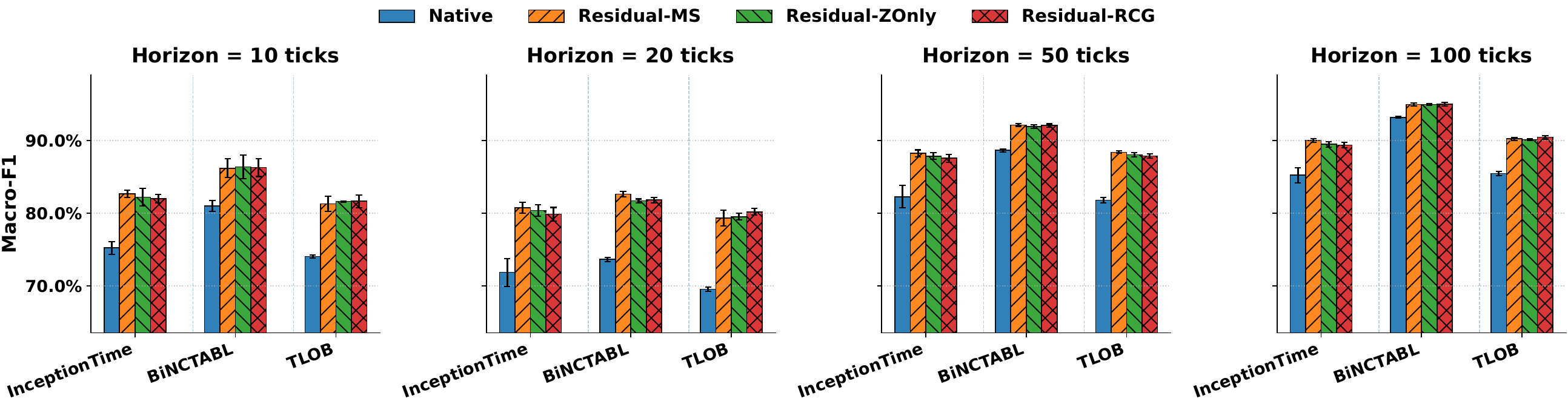}
    \caption{Residual multi-scale evidence improves FI-2010 forecasting across horizons, with the largest gains in short-horizon, high-noise regimes.}
    \label{fig:fi2010residual}
\end{figure}

\paragraph{Protocol.}
We evaluate the proposed decomposition across a diverse set of native temporal classifiers rather than relying on a single main backbone. For the general time-series benchmarks, the base models include convolutional, recurrent, residual, and attention-based architectures: FCN\citep{fcn}, GRU\citep{gru, gru2}, LSTM\citep{lstm}, ResNet1D\citep{resnet}, PatchTransformer\citep{patch}, and TemporalTransformer\citep{temporal}. For each base model, we attach three residual variants that isolate different forms of auxiliary temporal evidence: \emph{Residual-MS}, which uses uniform multi-scale fusion; \emph{Residual-ZOnly}, which learns scale trust directly from the learned scale representations; and \emph{Residual-RCG}, which conditions trust on both scale representations and explicit reliability descriptors. This design lets us test whether the representation--calibration decomposition is tied to a particular architecture or instead provides a more general layer that can be placed on top of heterogeneous temporal predictors.

On FI-2010, since the benchmark is specialized and relies on state-of-the-art models designed specifically for it, we apply the same decomposition to strong limit-order-book baselines, InceptionTime\citep{inceptiontime}, TLOB\citep{tlob} and BiNCTABL\citep{tran2021data}, in order to test whether residual multi-scale evidence remains useful in a high-frequency financial setting with specialized architectures. After training each raw model and its residual variants, we evaluate post-hoc calibration methods as diagnostic tools for the decision layer rather than as the central contribution. We distinguish raw residual gains, which indicate improved temporal evidence, from calibrated gains, which indicate whether native and auxiliary evidence can be converted into better final decisions. Complete details on the methodology of the experiments, training, hyperparameters, experimental setup, and reproducibility are provided in Appendix~\ref{app:experimental_setup}, while additional statistical results are reported in Appendix~\ref{app:additional_stats}.

\begin{figure}[t]
    \centering
    \includegraphics[width=0.9\textwidth]{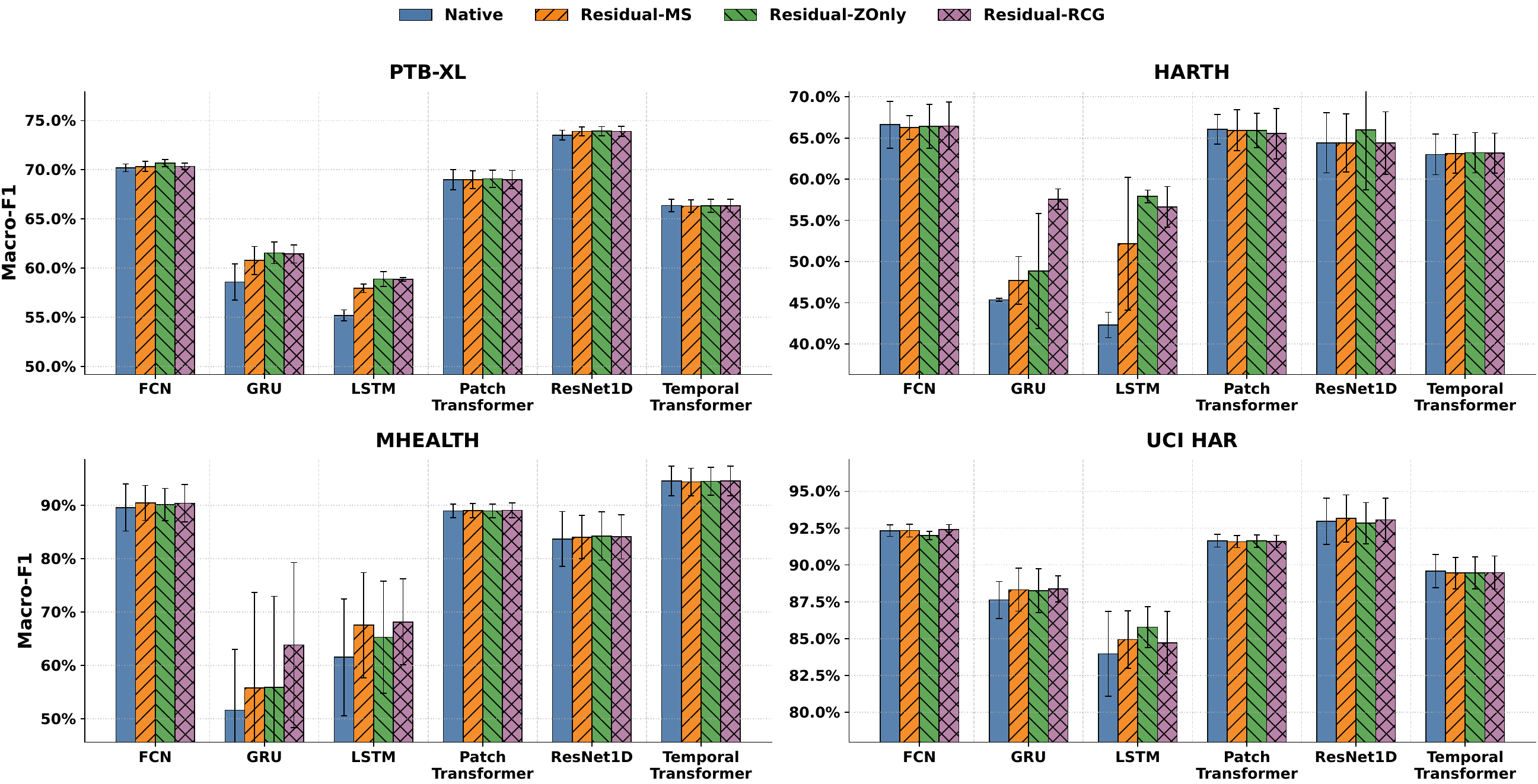}
    \caption{Residual gains are regime-dependent, concentrating in weaker recurrent backbones while stronger or near-saturated models show limited headroom.}
    \label{fig:datasetresidual}
\end{figure}

\paragraph{RQ1: When does residual evidence help?}
Residual evidence helps most when the native predictor is exposed to noisy or representation-limited temporal structure. This is clearest on FI-2010: across horizons, the residual multi-scale variants consistently improve over their native backbones, with the strongest effects in the short-horizon regime where limit-order-book dynamics are most local and unstable (Figure~\ref{fig:fi2010residual}). These results indicate that auxiliary multi-scale evidence is extremely useful when the native decision boundary alone is insufficient to resolve high-frequency ambiguity.

The broader benchmark results show a similar but more backbone-dependent pattern. Residual branches tend to help weaker recurrent models and produce smaller gains for stronger convolutional, residual, and Transformer-based predictors (Figure~\ref{fig:datasetresidual}). UCI-HAR is the main near-saturation case, where strong native performance leaves little room for residual correction. Overall, RQ1 supports the proposed decomposition: residual branches are most beneficial in noisy, under-resolved regimes, but become mostly neutral when the native representation is already stable. Full numerical results are reported in Appendix~\ref{app:additional_stats}.

\paragraph{RQ2: Does learned trust allocation improve over uniform fusion?}
Learned trust allocation improves over uniform fusion only selectively. On FI-2010, Figure~\ref{fig:fi2010residual} and Table~\ref{tab:native_adapters_macro_f1_horizons} show that Residual-MS is a strong baseline: it is the best raw residual variant in most horizon--backbone settings, including all InceptionTime horizons and all models at the 50-tick horizon. Residual-RCG is best in some TLOB and long-horizon settings, while Residual-ZOnly is rarely the clear winner. Thus, in the limit-order-book benchmark, adaptive trust is not consistently required once multi-scale residual evidence is available.

For the remaining benchmarks, Table~\ref{tab:dataset_residual_macro_f1} and Figure~\ref{fig:datasetresidual} show a more backbone-dependent pattern. Residual-ZOnly gives the strongest raw residual results for most PTB-XL backbones, indicating that learned scale representations are often sufficient for trust allocation. Residual-RCG is most useful in less stable recurrent activity-recognition settings, especially GRU on HARTH and GRU/LSTM on MHEALTH. In contrast, differences among MS, ZOnly, and RCG are small for stronger or near-saturated models, such as UCI-HAR and several Transformer or residual backbones.

Overall, learned trust allocation is useful but not uniformly superior to uniform fusion. ZOnly is often enough when scale representations already encode useful reliability information, while RCG helps mainly when explicit reliability descriptors capture instability not represented by the scale embeddings alone. The main conclusion is therefore regime-dependent: no trust-allocation rule dominates across datasets, backbones, and horizons.

\paragraph{RQ3: Do residual branches provide useful auxiliary evidence for post-hoc calibration?}
Yes, but only in regimes where the residual branch exposes evidence that the raw decision rule does not fully exploit. The clearest case is PTB-XL: Table~\ref{tab:appendix_best_calibration_compact} shows that recurrent backbones benefit more from calibrating residual models than from calibrating the native model alone, and the best methods are often branch-aware separated calibrators. This supports the interpretation that auxiliary residual logits contain decision-relevant information beyond the final raw logits. For stronger PTB-XL backbones, however, the calibration gains are small, indicating limited additional decision-level headroom.

\begin{wrapfigure}{r}{0.50\textwidth}
  \centering
  \vspace{-10pt}
  \includegraphics[width=0.44\textwidth]{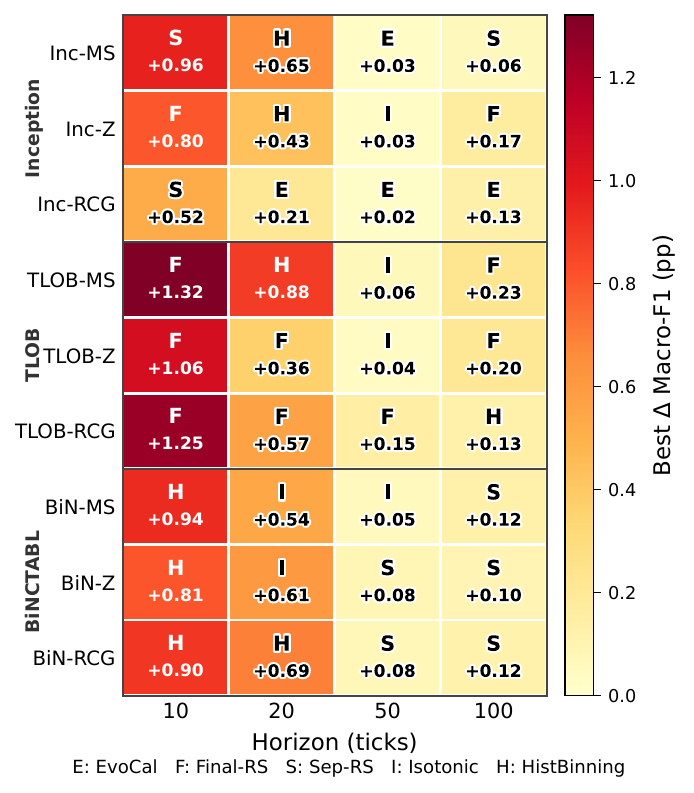}
  \caption{Best post-hoc calibration gain on FI-2010. Each cell reports the Macro-F1 gain, in percentage points, over the corresponding raw model. The letter indicates the calibration method that achieves the best gain. Full results in Appendix~\ref{app:additional_stats}.}
  \label{fig:fi2010_heatmap}
  \vspace{-10pt}
\end{wrapfigure}

This effect is not universal. On HARTH and MHEALTH, Table~\ref{tab:dataset_residual_macro_f1} and Table~\ref{tab:appendix_best_calibration_compact} show that the largest improvements coincide with large raw residual gains, so these cases are better interpreted as mixed representation--decision effects rather than calibration alone. FI-2010 shows a horizon-dependent pattern: Figure~\ref{fig:fi2010_heatmap} and Table~\ref{tab:appendix_fi2010_best_calibration_compact} indicate that calibration is most useful in shorter, noisier horizons and contributes little once the raw residual predictor is already strong. UCI-HAR shows the opposite regime, where near-saturated native models leave little room for either residual evidence or post-hoc recalibration.

\paragraph{RQ4: What regimes emerge across datasets?}
The results are better described by dataset--backbone regimes than by a single average effect. FI-2010 is the clearest residual-sensitive case: residual variants improve consistently across horizons and specialized backbones, while calibration adds smaller gains mainly in the shorter-horizon settings (Figures~\ref{fig:fi2010residual} and~\ref{fig:fi2010_heatmap}). In the remaining benchmarks, the effect depends more strongly on the native architecture. PTB-XL shows selective gains, concentrated in recurrent backbones and partly amplified by branch-aware calibration. HARTH and MHEALTH show larger residual opportunities for recurrent models, but weaker or neutral effects for several convolutional, residual, and Transformer-based backbones. UCI-HAR shows the smallest changes across native, residual, and calibrated variants, consistent with limited remaining headroom (Figure~\ref{fig:datasetresidual}).

Overall, the decomposition separates three empirical regimes. The first is residual-sensitive, where auxiliary temporal evidence changes the raw predictor, as in FI-2010 and recurrent-backbone settings on HARTH and MHEALTH. The second is calibration-sensitive, where branch-separated logits are useful mainly after decision-level recombination, as observed in parts of PTB-XL. The third is near-saturated, where neither residual correction nor calibration substantially changes performance, as in UCI-HAR and several stronger-backbone settings. Thus, the main conclusion is not that residuals or calibration dominate universally, but that the decomposition identifies which bottleneck is active for each dataset--backbone pair.

\paragraph{Summary.}
Overall, the results show that residual evidence, trust allocation, and post-hoc calibration are useful only in specific dataset--backbone regimes rather than as universal improvements. This supports the proposed decomposition as a diagnostic tool for distinguishing missing temporal evidence, underused decision-level evidence, and near-saturated settings.

\section{Conclusion}
\label{sec:conclusion}

We introduced a temporal classification framework based on a representation--calibration decomposition, separating native evidence from a frozen temporal backbone, conservative residual evidence from a multi-scale auxiliary branch, and post-hoc decision calibration over frozen logits. Across FI-2010, PTB-XL, HARTH, MHEALTH, and UCI-HAR, the results show that residual evidence and calibration are useful in different dataset--backbone regimes rather than as universal improvements. In harder and noisier settings, especially short-horizon FI-2010 and weaker recurrent backbones, residual branches improve the raw predictor by adding missing temporal evidence. In more stable settings, such as parts of PTB-XL, raw residual gains are smaller, but branch-aware calibration can still improve how native and auxiliary logits are converted into final decisions. In near-saturated regimes, such as UCI-HAR and several stronger-backbone settings, both interventions provide limited additional benefit. These findings support the paper's central claim: remaining errors in temporal classifiers are not uniformly representational, and explicitly separating evidence formation from decision correction provides a useful diagnostic framework for identifying whether a model is representation-limited, decision-limited, or close to saturation. The main limitations of our work and future directions are discussed in Appendix~\ref{app:limitations}.

\newpage

\bibliographystyle{plain}
\bibliography{refs}

@article{dempster2019rocket,
  title={ROCKET: exceptionally fast and accurate time series classification using random convolutional kernels},
  author={Dempster, Angus and Petitjean, Fran{\c{c}}ois and Webb, Geoffrey I},
  journal={arXiv preprint arXiv:1910.13051},
  year={2019}
}

@inproceedings{NIPS2017_3f5ee243,
 author = {Vaswani, Ashish and Shazeer, Noam and Parmar, Niki and Uszkoreit, Jakob and Jones, Llion and Gomez, Aidan N and Kaiser, \L ukasz and Polosukhin, Illia},
 booktitle = {Advances in Neural Information Processing Systems},
 editor = {I. Guyon and U. Von Luxburg and S. Bengio and H. Wallach and R. Fergus and S. Vishwanathan and R. Garnett},
 pages = {},
 publisher = {Curran Associates, Inc.},
 title = {Attention is All you Need},
 url = {https://proceedings.neurips.cc/paper_files/paper/2017/file/3f5ee243547dee91fbd053c1c4a845aa-Paper.pdf},
 volume = {30},
 year = {2017}
}

@inproceedings{NEURIPS2021_bcc0d400,
 author = {Wu, Haixu and Xu, Jiehui and Wang, Jianmin and Long, Mingsheng},
 booktitle = {Advances in Neural Information Processing Systems},
 editor = {M. Ranzato and A. Beygelzimer and Y. Dauphin and P.S. Liang and J. Wortman Vaughan},
 pages = {22419--22430},
 publisher = {Curran Associates, Inc.},
 title = {Autoformer: Decomposition Transformers with Auto-Correlation for Long-Term Series Forecasting},
 url = {https://proceedings.neurips.cc/paper_files/paper/2021/file/bcc0d400288793e8bdcd7c19a8ac0c2b-Paper.pdf},
 volume = {34},
 year = {2021}
}

@article{LIM20211748,
title = {Temporal Fusion Transformers for interpretable multi-horizon time series forecasting},
journal = {International Journal of Forecasting},
volume = {37},
number = {4},
pages = {1748-1764},
year = {2021},
issn = {0169-2070},
doi = {https://doi.org/10.1016/j.ijforecast.2021.03.012},
url = {https://www.sciencedirect.com/science/article/pii/S0169207021000637},
author = {Bryan Lim and Sercan Ö. Arık and Nicolas Loeff and Tomas Pfister}
}

@InProceedings{pmlr-v70-guo17a,
  title = 	 {On Calibration of Modern Neural Networks},
  author =       {Chuan Guo and Geoff Pleiss and Yu Sun and Kilian Q. Weinberger},
  booktitle = 	 {Proceedings of the 34th International Conference on Machine Learning},
  pages = 	 {1321--1330},
  year = 	 {2017},
  editor = 	 {Precup, Doina and Teh, Yee Whye},
  volume = 	 {70},
  series = 	 {Proceedings of Machine Learning Research},
  month = 	 {06--11 Aug},
  publisher =    {PMLR},
  url = 	 {https://proceedings.mlr.press/v70/guo17a.html}
}

@inproceedings{NEURIPS2019_8ca01ea9,
 author = {Kull, Meelis and Perello Nieto, Miquel and K\"{a}ngsepp, Markus and Silva Filho, Telmo and Song, Hao and Flach, Peter},
 booktitle = {Advances in Neural Information Processing Systems},
 editor = {H. Wallach and H. Larochelle and A. Beygelzimer and F. d\textquotesingle Alch\'{e}-Buc and E. Fox and R. Garnett},
 pages = {},
 publisher = {Curran Associates, Inc.},
 title = {Beyond temperature scaling: Obtaining well-calibrated multi-class probabilities with Dirichlet calibration},
 url = {https://proceedings.neurips.cc/paper_files/paper/2019/file/8ca01ea920679a0fe3728441494041b9-Paper.pdf},
 volume = {32},
 year = {2019}
}

@inproceedings{NEURIPS2019_f8c0c968,
 author = {Kumar, Ananya and Liang, Percy S and Ma, Tengyu},
 booktitle = {Advances in Neural Information Processing Systems},
 editor = {H. Wallach and H. Larochelle and A. Beygelzimer and F. d\textquotesingle Alch\'{e}-Buc and E. Fox and R. Garnett},
 pages = {},
 publisher = {Curran Associates, Inc.},
 title = {Verified Uncertainty Calibration},
 url = {https://proceedings.neurips.cc/paper_files/paper/2019/file/f8c0c968632845cd133308b1a494967f-Paper.pdf},
 volume = {32},
 year = {2019}
}

@InProceedings{10.1007/978-3-662-44851-9_15,
author="Lipton, Zachary C.
and Elkan, Charles
and Naryanaswamy, Balakrishnan",
editor="Calders, Toon
and Esposito, Floriana
and H{\"u}llermeier, Eyke
and Meo, Rosa",
title="Optimal Thresholding of Classifiers to Maximize F1 Measure",
booktitle="Machine Learning and Knowledge Discovery in Databases",
year="2014",
publisher="Springer Berlin Heidelberg",
address="Berlin, Heidelberg",
pages="225--239",
isbn="978-3-662-44851-9"
}

@misc{bai2018empiricalevaluationgenericconvolutional,
      title={An Empirical Evaluation of Generic Convolutional and Recurrent Networks for Sequence Modeling}, 
      author={Shaojie Bai and J. Zico Kolter and Vladlen Koltun},
      year={2018},
      eprint={1803.01271},
      archivePrefix={arXiv},
      primaryClass={cs.LG},
      url={https://arxiv.org/abs/1803.01271}, 
}

@inproceedings{yue2022ts2vec,
  title={Ts2vec: Towards universal representation of time series},
  author={Yue, Zhihan and Wang, Yujing and Duan, Juanyong and Yang, Tianmeng and Huang, Congrui and Tong, Yunhai and Xu, Bixiong},
  booktitle={Proceedings of the AAAI conference on artificial intelligence},
  volume={36},
  number={8},
  pages={8980--8987},
  year={2022}
}

@inproceedings{NEURIPS2022_266983d0,
  title = {{SCINet}: Time Series Modeling and Forecasting with Sample Convolution and Interaction},
  author = {Liu, Minhao and Zeng, Ailing and Chen, Muxi and Xu, Zhijian and Lai, Qiuxia and Ma, Lingna and Xu, Qiang},
  booktitle = {Advances in Neural Information Processing Systems},
  volume = {35},
  year = {2022},
  url = {https://proceedings.neurips.cc/paper_files/paper/2022/hash/266983d0949aed78a16fa4782237dea7-Abstract-Conference.html}
}

@article{wu2022timesnet,
  title={Timesnet: Temporal 2d-variation modeling for general time series analysis},
  author={Wu, Haixu and Hu, Tengge and Liu, Yong and Zhou, Hang and Wang, Jianmin and Long, Mingsheng},
  journal={arXiv preprint arXiv:2210.02186},
  year={2022}
}

@article{liu2023itransformer,
  title={itransformer: Inverted transformers are effective for time series forecasting},
  author={Liu, Yong and Hu, Tengge and Zhang, Haoran and Wu, Haixu and Wang, Shiyu and Ma, Lintao and Long, Mingsheng},
  journal={arXiv preprint arXiv:2310.06625},
  year={2023}
}

@inproceedings{NIPS2017_e7b24b11,
 author = {Rebuffi, Sylvestre-Alvise and Bilen, Hakan and Vedaldi, Andrea},
 booktitle = {Advances in Neural Information Processing Systems},
 editor = {I. Guyon and U. Von Luxburg and S. Bengio and H. Wallach and R. Fergus and S. Vishwanathan and R. Garnett},
 pages = {},
 publisher = {Curran Associates, Inc.},
 title = {Learning multiple visual domains with residual adapters},
 url = {https://proceedings.neurips.cc/paper_files/paper/2017/file/e7b24b112a44fdd9ee93bdf998c6ca0e-Paper.pdf},
 volume = {30},
 year = {2017}
}

@InProceedings{pmlr-v97-houlsby19a,
  title = 	 {Parameter-Efficient Transfer Learning for {NLP}},
  author =       {Houlsby, Neil and Giurgiu, Andrei and Jastrzebski, Stanislaw and Morrone, Bruna and De Laroussilhe, Quentin and Gesmundo, Andrea and Attariyan, Mona and Gelly, Sylvain},
  booktitle = 	 {Proceedings of the 36th International Conference on Machine Learning},
  pages = 	 {2790--2799},
  year = 	 {2019},
  editor = 	 {Chaudhuri, Kamalika and Salakhutdinov, Ruslan},
  volume = 	 {97},
  series = 	 {Proceedings of Machine Learning Research},
  month = 	 {09--15 Jun},
  publisher =    {PMLR},
  url = 	 {https://proceedings.mlr.press/v97/houlsby19a.html}
}

@article{lakshminarayanan2017simple,
  title={Simple and scalable predictive uncertainty estimation using deep ensembles},
  author={Lakshminarayanan, Balaji and Pritzel, Alexander and Blundell, Charles},
  journal={Advances in neural information processing systems},
  volume={30},
  year={2017}
}

@inproceedings{10.1145/775047.775151,
author = {Zadrozny, Bianca and Elkan, Charles},
title = {Transforming classifier scores into accurate multiclass probability estimates},
year = {2002},
isbn = {158113567X},
publisher = {Association for Computing Machinery},
address = {New York, NY, USA},
url = {https://doi.org/10.1145/775047.775151},
doi = {10.1145/775047.775151},
booktitle = {Proceedings of the Eighth ACM SIGKDD International Conference on Knowledge Discovery and Data Mining},
pages = {694–699},
numpages = {6},
location = {Edmonton, Alberta, Canada},
series = {KDD '02}
}

@inproceedings{10.1145/1102351.1102430,
author = {Niculescu-Mizil, Alexandru and Caruana, Rich},
title = {Predicting good probabilities with supervised learning},
year = {2005},
isbn = {1595931805},
publisher = {Association for Computing Machinery},
address = {New York, NY, USA},
url = {https://doi.org/10.1145/1102351.1102430},
doi = {10.1145/1102351.1102430},
booktitle = {Proceedings of the 22nd International Conference on Machine Learning},
pages = {625–632},
numpages = {8},
location = {Bonn, Germany},
series = {ICML '05}
}

@article{zhang2022memo,
  title={Memo: Test time robustness via adaptation and augmentation},
  author={Zhang, Marvin and Levine, Sergey and Finn, Chelsea},
  journal={Advances in neural information processing systems},
  volume={35},
  pages={38629--38642},
  year={2022}
}

@article{zhang2019deeplob,
  title={Deeplob: Deep convolutional neural networks for limit order books},
  author={Zhang, Zihao and Zohren, Stefan and Roberts, Stephen},
  journal={IEEE Transactions on Signal Processing},
  volume={67},
  number={11},
  pages={3001--3012},
  year={2019},
  publisher={IEEE}
}

@inproceedings{tran2021data,
  title={Data normalization for bilinear structures in high-frequency financial time-series},
  author={Tran, Dat Thanh and Kanniainen, Juho and Gabbouj, Moncef and Iosifidis, Alexandros},
  booktitle={2020 25th International conference on pattern recognition (ICPR)},
  pages={7287--7292},
  year={2021},
  organization={IEEE}
}

@misc{human_activity_recognition_using_smartphones_240,
  author       = {Reyes-Ortiz, Jorge and Anguita, Davide and Ghio, Alessandro and Oneto, Luca and Parra, Xavier},
  title        = {{Human Activity Recognition Using Smartphones}},
  year         = {2013},
  howpublished = {UCI Machine Learning Repository},
  note         = {{DOI}: https://doi.org/10.24432/C54S4K}
}

@misc{mhealth_319,
  author       = {Banos, Oresti and Garcia, Rafael and Saez, Alejandro},
  title        = {{MHEALTH}},
  year         = {2014},
  howpublished = {UCI Machine Learning Repository},
  note         = {{DOI}: https://doi.org/10.24432/C5TW22}
}

@article{fi2010,
author = {Ntakaris, Adamantios and Magris, Martin and Kanniainen, Juho and Gabbouj, Moncef and Iosifidis, Alexandros},
title = {Benchmark dataset for mid-price forecasting of limit order book data with machine learning methods},
journal = {Journal of Forecasting},
volume = {37},
number = {8},
pages = {852-866},
doi = {https://doi.org/10.1002/for.2543},
url = {https://onlinelibrary.wiley.com/doi/abs/10.1002/for.2543},
eprint = {https://onlinelibrary.wiley.com/doi/pdf/10.1002/for.2543},
year = {2018}
}

@Article{ptbxl,
author={Wagner, Patrick
and Strodthoff, Nils
and Bousseljot, Ralf-Dieter
and Kreiseler, Dieter
and Lunze, Fatima I.
and Samek, Wojciech
and Schaeffter, Tobias},
title={PTB-XL, a large publicly available electrocardiography dataset},
journal={Scientific Data},
year={2020},
month={May},
day={25},
volume={7},
number={1},
pages={154},
issn={2052-4463},
doi={10.1038/s41597-020-0495-6},
url={https://doi.org/10.1038/s41597-020-0495-6}
}

@Article{harth,
AUTHOR = {Logacjov, Aleksej and Bach, Kerstin and Kongsvold, Atle and Bårdstu, Hilde Bremseth and Mork, Paul Jarle},
TITLE = {HARTH: A Human Activity Recognition Dataset for Machine Learning},
JOURNAL = {Sensors},
VOLUME = {21},
YEAR = {2021},
NUMBER = {23},
ARTICLE-NUMBER = {7853},
URL = {https://www.mdpi.com/1424-8220/21/23/7853},
PubMedID = {34883863},
ISSN = {1424-8220},
DOI = {10.3390/s21237853}
}

@InProceedings{mhealth,
author="Banos, Oresti
and Garcia, Rafael
and Holgado-Terriza, Juan A.
and Damas, Miguel
and Pomares, Hector
and Rojas, Ignacio
and Saez, Alejandro
and Villalonga, Claudia",
editor="Pecchia, Leandro
and Chen, Liming Luke
and Nugent, Chris
and Bravo, Jos{\'e}",
title="mHealthDroid: A Novel Framework for Agile Development of Mobile Health Applications",
booktitle="Ambient Assisted Living and Daily Activities",
year="2014",
publisher="Springer International Publishing",
address="Cham",
pages="91--98",
isbn="978-3-319-13105-4"
}

@inproceedings{uci_har,
  title={A Public Domain Dataset for Human Activity Recognition using Smartphones},
  author={D. Anguita and Alessandro Ghio and L. Oneto and Xavier Parra and Jorge Luis Reyes-Ortiz},
  booktitle={The European Symposium on Artificial Neural Networks},
  year={2013},
  url={https://api.semanticscholar.org/CorpusID:6975432}
}

@article{tlob,
  title={TLOB: A Novel Transformer Model with Dual Attention for Stock Price Trend Prediction with Limit Order Book Data},
  author={Berti, Leonardo and Kasneci, Gjergji},
  journal={arXiv preprint arXiv:2502.15757},
  year={2025}
}

@article{inceptiontime,
author = {Ismail Fawaz, Hassan and Lucas, Benjamin and Forestier, Germain and Pelletier, Charlotte and Schmidt, Daniel F. and Weber, Jonathan and Webb, Geoffrey I. and Idoumghar, Lhassane and Muller, Pierre-Alain and Petitjean, Fran\c{c}ois},
title = {InceptionTime: Finding AlexNet for time series classification},
year = {2020},
issue_date = {Nov 2020},
publisher = {Kluwer Academic Publishers},
address = {USA},
volume = {34},
number = {6},
issn = {1384-5810},
url = {https://doi.org/10.1007/s10618-020-00710-y},
doi = {10.1007/s10618-020-00710-y},
journal = {Data Min. Knowl. Discov.},
month = nov,
pages = {1936–1962},
numpages = {27}
}

@article{fcn,
  title={Time series classification from scratch with deep neural networks: A strong baseline},
  author={Zhiguang Wang and Weizhong Yan and Tim Oates},
  journal={2017 International Joint Conference on Neural Networks (IJCNN)},
  year={2017},
  pages={1578-1585},
  url={https://api.semanticscholar.org/CorpusID:14303613}
}

@inproceedings{gru,
    title = "Learning Phrase Representations using {RNN} Encoder{--}Decoder for Statistical Machine Translation",
    author = {Cho, Kyunghyun  and
      van Merri{\"e}nboer, Bart  and
      Gulcehre, Caglar  and
      Bahdanau, Dzmitry  and
      Bougares, Fethi  and
      Schwenk, Holger  and
      Bengio, Yoshua},
    editor = "Moschitti, Alessandro  and
      Pang, Bo  and
      Daelemans, Walter",
    booktitle = "Proceedings of the 2014 Conference on Empirical Methods in Natural Language Processing ({EMNLP})",
    month = oct,
    year = "2014",
    address = "Doha, Qatar",
    publisher = "Association for Computational Linguistics",
    url = "https://aclanthology.org/D14-1179/",
    doi = "10.3115/v1/D14-1179",
    pages = "1724--1734"
}

@article{lstm,
author = {Hochreiter, Sepp and Schmidhuber, J\"{u}rgen},
title = {Long Short-Term Memory},
year = {1997},
issue_date = {November 15, 1997},
publisher = {MIT Press},
address = {Cambridge, MA, USA},
volume = {9},
number = {8},
issn = {0899-7667},
url = {https://doi.org/10.1162/neco.1997.9.8.1735},
doi = {10.1162/neco.1997.9.8.1735},
journal = {Neural Comput.},
month = nov,
pages = {1735–1780},
numpages = {46}
}

@article{resnet,
  title={Deep Residual Learning for Image Recognition},
  author={Kaiming He and X. Zhang and Shaoqing Ren and Jian Sun},
  journal={2016 IEEE Conference on Computer Vision and Pattern Recognition (CVPR)},
  year={2015},
  pages={770-778},
  url={https://api.semanticscholar.org/CorpusID:206594692}
}

@article{gru2,
  title={Empirical Evaluation of Gated Recurrent Neural Networks on Sequence Modeling},
  author={Junyoung Chung and Çaglar G{\"u}lçehre and Kyunghyun Cho and Yoshua Bengio},
  journal={ArXiv},
  year={2014},
  volume={abs/1412.3555},
  url={https://api.semanticscholar.org/CorpusID:5201925}
}

@inproceedings{
patch,
title={A Time Series is Worth 64 Words:  Long-term Forecasting with Transformers},
author={Yuqi Nie and Nam H Nguyen and Phanwadee Sinthong and Jayant Kalagnanam},
booktitle={The Eleventh International Conference on Learning Representations },
year={2023},
url={https://openreview.net/forum?id=Jbdc0vTOcol}
}

@inproceedings{temporal,
author = {Zerveas, George and Jayaraman, Srideepika and Patel, Dhaval and Bhamidipaty, Anuradha and Eickhoff, Carsten},
title = {A Transformer-based Framework for Multivariate Time Series Representation Learning},
year = {2021},
isbn = {9781450383325},
publisher = {Association for Computing Machinery},
address = {New York, NY, USA},
url = {https://doi.org/10.1145/3447548.3467401},
doi = {10.1145/3447548.3467401},
booktitle = {Proceedings of the 27th ACM SIGKDD Conference on Knowledge Discovery \& Data Mining},
pages = {2114–2124},
numpages = {11},
location = {Virtual Event, Singapore},
series = {KDD '21}
}

@article{bagnall,
author = {Bagnall, Anthony and Lines, Jason and Bostrom, Aaron and Large, James and Keogh, Eamonn},
title = {The great time series classification bake off: a review and experimental evaluation of recent algorithmic advances},
year = {2017},
issue_date = {May       2017},
publisher = {Kluwer Academic Publishers},
address = {USA},
volume = {31},
number = {3},
issn = {1384-5810},
url = {https://doi.org/10.1007/s10618-016-0483-9},
doi = {10.1007/s10618-016-0483-9},
journal = {Data Min. Knowl. Discov.},
month = may,
pages = {606–660},
numpages = {55}
}

@article{ruizetal,
author = {Ruiz, Alejandro Pasos and Flynn, Michael and Large, James and Middlehurst, Matthew and Bagnall, Anthony},
title = {The great multivariate time series classification bake off: a review and experimental evaluation of recent algorithmic advances},
year = {2021},
issue_date = {Mar 2021},
publisher = {Kluwer Academic Publishers},
address = {USA},
volume = {35},
number = {2},
issn = {1384-5810},
url = {https://doi.org/10.1007/s10618-020-00727-3},
doi = {10.1007/s10618-020-00727-3},
journal = {Data Min. Knowl. Discov.},
month = mar,
pages = {401–449},
numpages = {49}
}

@article{hivecote,
author = {Middlehurst, Matthew and Large, James and Flynn, Michael and Lines, Jason and Bostrom, Aaron and Bagnall, Anthony},
title = {HIVE-COTE 2.0: a new meta ensemble for time series classification},
year = {2021},
issue_date = {Dec 2021},
publisher = {Kluwer Academic Publishers},
address = {USA},
volume = {110},
number = {11–12},
issn = {0885-6125},
url = {https://doi.org/10.1007/s10994-021-06057-9},
doi = {10.1007/s10994-021-06057-9},
journal = {Mach. Learn.},
month = dec,
pages = {3211–3243},
numpages = {33}
}

@Article{Tan2022,
author={Tan, Chang Wei
and Dempster, Angus
and Bergmeir, Christoph
and Webb, Geoffrey I.},
title={MultiRocket: multiple pooling operators and transformations for fast and effective time series classification},
journal={Data Mining and Knowledge Discovery},
year={2022},
month={Sep},
day={01},
volume={36},
number={5},
pages={1623-1646},
issn={1573-756X},
doi={10.1007/s10618-022-00844-1},
url={https://doi.org/10.1007/s10618-022-00844-1}
}

@inproceedings{Vaicenavicius2019EvaluatingMC,
  title={Evaluating model calibration in classification},
  author={Juozas Vaicenavicius and David Widmann and Carl R. Andersson and Fredrik Lindsten and Jacob Roll and Thomas Bo Sch{\"o}n},
  booktitle={International Conference on Artificial Intelligence and Statistics},
  year={2019},
  url={https://api.semanticscholar.org/CorpusID:67749814}
}

@inproceedings{minderer,
author = {Minderer, Matthias and Djolonga, Josip and Romijnders, Rob and Hubis, Frances and Zhai, Xiaohua and Houlsby, Neil and Tran, Dustin and Lucic, Mario},
title = {Revisiting the calibration of modern neural networks},
year = {2021},
isbn = {9781713845393},
publisher = {Curran Associates Inc.},
address = {Red Hook, NY, USA},
booktitle = {Proceedings of the 35th International Conference on Neural Information Processing Systems},
articleno = {1200},
numpages = {13},
series = {NIPS '21}
}

@inproceedings{yenan,
author = {Ye, Nan and Chai, Kian Ming A. and Lee, Wee Sun and Chieu, Hai Leong},
title = {Optimizing F-measures: a tale of two approaches},
year = {2012},
isbn = {9781450312851},
publisher = {Omnipress},
address = {Madison, WI, USA},
booktitle = {Proceedings of the 29th International Coference on International Conference on Machine Learning},
pages = {1555–1562},
numpages = {8},
location = {Edinburgh, Scotland},
series = {ICML'12}
}

@inproceedings{parambath,
 author = {Parambath, Shameem A. Puthiya and Usunier, Nicolas and Grandvalet, Yves},
 booktitle = {Advances in Neural Information Processing Systems},
 editor = {Z. Ghahramani and M. Welling and C. Cortes and N. Lawrence and K. Weinberger},
 pages = {},
 publisher = {Curran Associates, Inc.},
 title = {Optimizing F-Measures by Cost-Sensitive Classification},
 url = {https://proceedings.neurips.cc/paper_files/paper/2014/file/5c0314ec1b57fcd36bbb013f3f025868-Paper.pdf},
 volume = {27},
 year = {2014}
}

@InProceedings{dembczynski,
  title = 	 {Optimizing the F-Measure in Multi-Label Classification: Plug-in Rule Approach versus Structured Loss Minimization},
  author = 	 {Dembczynski, Krzysztof and Jachnik, Arkadiusz and Kotlowski, Wojciech and Waegeman, Willem and Huellermeier, Eyke},
  booktitle = 	 {Proceedings of the 30th International Conference on Machine Learning},
  pages = 	 {1130--1138},
  year = 	 {2013},
  editor = 	 {Dasgupta, Sanjoy and McAllester, David},
  volume = 	 {28},
  number =       {3},
  series = 	 {Proceedings of Machine Learning Research},
  address = 	 {Atlanta, Georgia, USA},
  month = 	 {17--19 Jun},
  publisher =    {PMLR},
  url = 	 {https://proceedings.mlr.press/v28/dembczynski13.html}
}

@inproceedings{yusun,
author = {Sun, Yu and Wang, Xiaolong and Liu, Zhuang and Miller, John and Efros, Alexei A. and Hardt, Moritz},
title = {Test-time training with self-supervision for generalization under distribution shifts},
year = {2020},
publisher = {JMLR.org},
booktitle = {Proceedings of the 37th International Conference on Machine Learning},
articleno = {856},
numpages = {20},
series = {ICML'20}
}

@inproceedings{
wang2021tent,
title={Tent: Fully Test-Time Adaptation by Entropy Minimization},
author={Dequan Wang and Evan Shelhamer and Shaoteng Liu and Bruno Olshausen and Trevor Darrell},
booktitle={International Conference on Learning Representations},
year={2021},
url={https://openreview.net/forum?id=uXl3bZLkr3c}
}

\newpage
\appendix
\counterwithin{table}{section}
\counterwithin{figure}{section}
\renewcommand{\thetable}{\thesection.\arabic{table}}
\renewcommand{\thefigure}{\thesection.\arabic{figure}}

\section{Additional Method Details and Reproducibility}
\label{app:experimental_setup}

This appendix provides the mathematical and implementation details required to reproduce the experiments. The central design choice is to train the native predictor first and then freeze it. Residual branches are trained afterwards as conservative additive corrections on top of frozen native logits. Post-hoc calibration is then fitted only on validation logits, without updating any representation parameters.

All code, data-processing instructions, hyperparameters, saved artifacts, and running instructions are described in the Supplementary Material.

\subsection{Native Predictor}

Let \(x\in\mathbb{R}^{T\times D}\) be a multivariate temporal input with \(T\) time steps and \(D\) channels. A native temporal classifier \(f_{\psi}\) produces logits
\(
\ell_{\mathrm{nat}}(x)
=
f_{\psi}(x)
\in\mathbb{R}^{K},
\)
where \(K\) is the number of output classes. The native model is trained first using the task loss on the training split and selected by validation macro-F1 through early stopping. After this stage, all native parameters \(\psi\) are frozen.

For single-label datasets, the native objective is weighted cross-entropy:
\[
\mathcal{L}_{\mathrm{CE}}
=
-\frac{1}{N}
\sum_{i=1}^{N}
w_{y_i}
\log
\frac{
\exp(\ell_{\mathrm{nat},y_i}(x_i))
}{
\sum_{k=1}^{K}
\exp(\ell_{\mathrm{nat},k}(x_i))
},
\]
where class weights are inverse-frequency weights normalized to have mean one. For PTB-XL, which is multi-label, the corresponding binary cross-entropy with logits is used.

\subsection{Residual Multi-Scale Adapter}

Given a frozen native classifier, the residual adapter constructs auxiliary temporal evidence from multiple temporal scales. We use the scale set
\(
\mathcal{S}=\{1,2,4\},
\space
M=|\mathcal{S}|=3.
\)
For each scale \(s\in\mathcal{S}\), the input is downsampled by average pooling when \(s>1\):
\[
x^{(s)}
=
\begin{cases}
x, & s=1,\\
\operatorname{AvgPool}_{s}(x), & s>1.
\end{cases}
\]
Each scale is encoded by an auxiliary encoder \(E_s\), producing hidden states
\(
H^{(s)}
=
E_s(x^{(s)})
\in\mathbb{R}^{T_s\times d},
\)
with \(d=64\). A scale representation
\(
z^{(s)}\in\mathbb{R}^{d}
\)
is obtained either by attention pooling in the InceptionTime/FI-2010 residual adapter or by temporal averaging in the generic cross-backbone adapter.

\paragraph{Cross-scale representation attention.}
In the InceptionTime and FI-2010 residual adapters, the scale representations are stacked as
\(
Z
=
[z^{(s_1)},\dots,z^{(s_M)}]
\in\mathbb{R}^{M\times d}.
\)
A cross-scale attention layer refines these scale representations:
\[
Q_{\mathrm{sc}}=ZW_Q,
\qquad
K_{\mathrm{sc}}=ZW_K,
\qquad
V_{\mathrm{sc}}=ZW_V,
\]
\[
A_{\mathrm{sc}}
=
\softmax\!\left(
\frac{
Q_{\mathrm{sc}}K_{\mathrm{sc}}^\top
}{
\sqrt{d}
}
\right),
\qquad
\widetilde{Z}
=
\LN(Z+A_{\mathrm{sc}}V_{\mathrm{sc}}).
\]
The refined representation at scale \(s\) is denoted \(\tilde z^{(s)}\). In the generic cross-backbone adapter used for FCN, GRU, LSTM, ResNet1D, PatchTransformer, and TemporalTransformer, this cross-scale attention block is omitted and the scale representations are used directly, so \(\tilde z^{(s)}=z^{(s)}\).

\subsection{Scale Trust Allocation}

The residual branch assigns sample-wise trust weights over scales. These weights satisfy
\(
\alpha_s(x)\ge 0,
\space
\sum_{s\in\mathcal{S}}\alpha_s(x)=1.
\)

\paragraph{Residual-MS.}
Residual-MS uses uniform scale trust:
\(
\alpha_s(x)
=
\frac{1}{M},
\space
\forall s\in\mathcal{S}.
\)

\paragraph{Residual-ZOnly.}
Residual-ZOnly learns scale trust from scale representations alone. In the InceptionTime/FI-2010 adapter, each scale receives a score
\(
a_s(x)
=
g_{\eta}(\tilde z^{(s)}(x)),
\)
and the adaptive weights are
\[
\alpha^{\mathrm{ad}}_s(x)
=
\frac{\exp(a_s(x)/\tau)}
{\sum_{r\in\mathcal{S}}\exp(a_r(x)/\tau)},
\qquad
\tau=1.5.
\]
The final weights interpolate between uniform and adaptive trust:
\[
\alpha_s(x)
=
(1-\mu)\frac{1}{M}
+
\mu\,\alpha^{\mathrm{ad}}_s(x),
\qquad
\mu=\sigma(\eta_{\mu}),
\]
where \(\eta_{\mu}\) is a learned scalar. In the generic cross-backbone adapter, the gate receives the concatenated scale representations and directly outputs a softmax distribution over scales.

\paragraph{Residual-RCG.}
Residual-RCG augments scale representations with explicit reliability descriptors. In the InceptionTime/FI-2010 adapter, for each scale \(s\), the descriptor vector is
\[
r^{(s)}(x)
=
\big(
e^{(s)}(x),
u^{(s)}(x),
c^{(s)}(x),
\nu^{(s)}(x)
\big),
\]
where
\[
e^{(s)}(x)
=
\frac{1}{(T_s-1)d}
\sum_{t=2}^{T_s}
\left\|
H^{(s)}_{t}
-
H^{(s)}_{t-1}
\right\|_2^2
\]
measures local temporal instability,
\[
u^{(s)}(x)
=
\frac{1}{w_s d}
\sum_{t=T_s-w_s+1}^{T_s}
\left\|
H^{(s)}_{t}
-
\frac{1}{w_s}
\sum_{\tau=T_s-w_s+1}^{T_s}
H^{(s)}_{\tau}
\right\|_2^2
\]
measures recent instability over the last \(w_s=\min(8,T_s)\) hidden states,
\[
c^{(s)}(x)
=
-\frac{1}{|\mathcal{N}(s)|}
\sum_{r\in\mathcal{N}(s)}
\frac{1}{d}
\left\|
\tilde z^{(s)}(x)
-
\tilde z^{(r)}(x)
\right\|_2^2
\]
measures consistency with neighboring scales, and
\[
\nu^{(s)}(x)
=
\frac{1}{d}
\left\|
\tilde z^{(s)}(x)
\right\|_2^2
\]
measures representation energy. Here \(\mathcal{N}(s)\) denotes the adjacent scales of \(s\). The descriptor matrix is standardized within each batch and clipped to the interval \([-5,5]\). Descriptor ablations remove one component at a time: NoE removes \(e\), NoU removes \(u\), NoC removes \(c\), and NoNu removes \(\nu\).

The RCG gate computes
\(
a_s(x)
=
g_{\eta}\big([\tilde z^{(s)}(x);r^{(s)}(x)]\big),
\)
followed by the same temperature-softmax and uniform-adaptive mixture as in Residual-ZOnly.

In the generic cross-backbone adapter, RCG uses a simpler raw-input descriptor vector per scale:
\[
r^{(s)}_{\mathrm{gen}}(x)
=
\big(
\operatorname{mean}((x^{(s)})^2),
\operatorname{mean}(|x^{(s)}|),
\operatorname{std}(x^{(s)}),
\operatorname{mean}(|\Delta_t x^{(s)}|)
\big),
\]
concatenated with the scale representations before the softmax gate.

\subsection{Residual Logits and Training Objective}

The auxiliary representation is the trust-weighted sum of scale representations:
\[
z_{\mathrm{aux}}(x)
=
\sum_{s\in\mathcal{S}}
\alpha_s(x)\tilde z^{(s)}(x),
\]
or the analogous sum over non-attended scale representations in the generic adapter. Auxiliary logits are then computed as
\(
\ell_{\mathrm{aux}}(x)
=
W_{\mathrm{aux}}z_{\mathrm{aux}}(x)+b_{\mathrm{aux}}.
\)
The final raw residual logits are
\[
\ell_{\mathrm{raw}}(x)
=
\ell_{\mathrm{nat}}(x)
+
\gamma_{\mathrm{res}}\ell_{\mathrm{aux}}(x),
\qquad
\gamma_{\mathrm{res}}=\sigma(\rho),
\]
where \(\rho\) is a learned scalar initialized to \(-3.0\). Thus, the residual branch begins as a small correction to the frozen native classifier.

For the InceptionTime and FI-2010 residual adapters, the residual objective is
\[
\mathcal{L}_{\mathrm{res}}
=
\mathcal{L}_{\mathrm{task}}(\ell_{\mathrm{raw}},y)
+
\lambda_{\mathrm{aux}}
\mathcal{L}_{\mathrm{task}}(\ell_{\mathrm{aux}},y)
+
\lambda_{\mathrm{cons}}
\mathcal{L}_{\mathrm{cons}}
+
\lambda_{\mathrm{ent}}
\mathcal{L}_{\mathrm{ent}},
\]
with
\(
\lambda_{\mathrm{aux}}=0.30,
\space
\lambda_{\mathrm{cons}}=0.005,
\space
\lambda_{\mathrm{ent}}=0.001.
\)

The consistency penalty is
\[
\mathcal{L}_{\mathrm{cons}}
=
\frac{1}{M-1}
\sum_{m=1}^{M-1}
\left\|
\tilde z^{(s_m)}(x)
-
\tilde z^{(s_{m+1})}(x)
\right\|_2^2,
\]
and the entropy regularization term is
\[
\mathcal{L}_{\mathrm{ent}}
=
\frac{1}{N}
\sum_{i=1}^{N}
\sum_{s\in\mathcal{S}}
\alpha_s(x_i)\log \alpha_s(x_i).
\]
In the generic cross-backbone residual adapter, the residual branch is optimized with the task loss on \(\ell_{\mathrm{raw}}\), while the native branch remains frozen.

\subsection{Post-hoc Decision Calibration}

After training a raw model, all model parameters are frozen. Calibration is fitted only on validation logits and then evaluated on test logits. For a residual model, we store both native and auxiliary logits, enabling branch-aware calibration.

Branch-agnostic calibration operates on \(\ell_{\mathrm{raw}}\). We evaluate scalar temperature scaling,
\(
\mathcal{C}(\ell)
=
\frac{\ell}{T},
\space T>0,
\)
vector scaling,
\(
\mathcal{C}(\ell)
=
\ell\odot a+b_{\mathrm{cal}},
\)
matrix scaling,
\(
\mathcal{C}(\ell)
=
A_{\mathrm{cal}}\ell+b_{\mathrm{cal}},
\)
and adaptive temperature scaling,
\[
\mathcal{C}(\ell)
=
\frac{\ell}{T(\ell)},
\qquad
T(\ell)
=
\exp\!\left(
a_{\mathrm{temp}}
+
b_{\mathrm{temp}}
\left(
1-\max_k \softmax(\ell)_k
\right)
\right).
\]
These differentiable calibrators are fitted with LBFGS for at most 300 iterations using the validation loss.

We also evaluate nonparametric calibration. Isotonic regression fits one isotonic model per class to map validation probabilities to calibrated class probabilities. Histogram binning uses 15 bins per class.

Search-based final-only calibration applies
\(
\mathcal{C}_{\mathrm{final}}(\ell)
=
\frac{\ell}{T}
+
b_{\mathrm{cal}},
\space T>0.
\)
The random-search version uses 800 trials. The temperature is sampled in the range \([0.35,5.0]\), and each bias coordinate is sampled in \([-3,3]\).

For residual models, branch-aware separated calibration preserves native and auxiliary evidence:
\[
\mathcal{C}_{\mathrm{sep}}
(\ell_{\mathrm{nat}},\ell_{\mathrm{aux}})
=
\frac{\ell_{\mathrm{nat}}}{T_{\mathrm{nat}}}
+
\Gamma\odot
\left(
\frac{\ell_{\mathrm{aux}}}{T_{\mathrm{aux}}}
\right)
+
b_{\mathrm{cal}},
\]
where \(T_{\mathrm{nat}}>0\) and \(T_{\mathrm{aux}}>0\) are positive temperatures, \(\Gamma\in[0,3]^K\) is a class-wise residual gain vector, and \(b_{\mathrm{cal}}\in\mathbb{R}^{K}\) is a calibration bias vector. This parameterization lets the calibrator adjust the strength of the native branch, the auxiliary residual branch, and class-specific decision biases without updating the representation.

We consider two search-based optimizers for this parameterization. RandomSearch-Separated samples 800 candidate calibrators from the predefined parameter ranges. EvoCal uses the same final-only or separated calibration forms, but optimizes their parameters with an evolutionary search. Each individual represents a complete calibrator, including temperatures, class-wise residual gains when applicable, and bias terms. Individuals are selected according to validation macro-F1, and new candidates are generated through elitism, random injection, and mutation. In our experiments, EvoCal uses population size 80, 80 generations, elite fraction 0.20, random fraction 0.15, mutation scale 0.10, temperature range \([0.35,5.0]\), gain range \([0,3]\), and bias range \([-3,3]\). We report EvoCal-F for the final-logit version and EvoCal-S for the branch-aware separated version.

Search-based calibrators optimize validation macro-F1 directly. Thus, unlike likelihood-based calibration, EvoCal and random-search calibration are designed to tune the inference-time decision rule for the same metric used in the main evaluation.

\subsection{Training Hyperparameters}

All non-FI-2010 experiments use seeds \(\{42,43,44\}\), batch size 128, AdamW optimizer, learning rate \(3\times 10^{-4}\), weight decay \(10^{-4}\), gradient clipping at 1.0, dropout 0.1, patience 4, and no dataloader workers. Native and baseline models are trained for at most 15 epochs; residual adapters are trained for at most 10 epochs. The residual branch uses \(d_{\mathrm{model}}=64\), scale factors \((1,2,4)\), temperature \(\tau=1.5\), reliability hidden size 64, residual-logit initialization \(\rho=-3.0\), consistency weight \(\lambda_{\mathrm{cons}}=0.005\), entropy weight \(\lambda_{\mathrm{ent}}=0.001\), and auxiliary task-loss weight \(\lambda_{\mathrm{aux}}=0.30\).

HARTH and MHEALTH use window size 250 and stride 125. HARTH uses split seed 123 with train/validation/test proportions determined by the experimental script, with 70\% of the development split used for training and 15\% for validation. MHEALTH uses split seed 456, drops the null class, and uses the same 70\%/15\% train/validation protocol. UCI-HAR uses all nine channels, validation subject fraction 0.20, and split seed 321. PTB-XL uses 100Hz ECG signals and superclass labels.

FI-2010 uses the same seeds \(\{42,43,44\}\), AdamW, learning rate \(3\times10^{-4}\), weight decay \(10^{-4}\), gradient clipping at 1.0, patience 4, dropout 0.1, \(d_{\mathrm{model}}=64\), scale factors \((1,2,4)\), temperature \(\tau=1.5\), reliability hidden size 64, residual-logit initialization \(\rho=-3.0\), consistency weight \(\lambda_{\mathrm{cons}}=0.005\), entropy weight \(\lambda_{\mathrm{ent}}=0.001\), and auxiliary task-loss weight \(\lambda_{\mathrm{aux}}=0.30\). The batch size is 256. Native and baseline models are trained for at most 15 epochs, and residual adapters for at most 10 epochs. FI-2010 additionally uses \(n_{\mathrm{head}}=4\), TLOB temporal layers \(=2\), TLOB variable layers \(=2\), and BiNCTABL depth \(=3\).

For FI-2010, inputs use the first 40 limit-order-book features. The sequence length is 100, and windows are formed within each split with the label assigned at the final time step of the window. Labels are normalized to \(\{0,1,2\}\), either from the original \(\{-1,0,1\}\) encoding or from the \(\{1,2,3\}\) encoding. The horizon map is
\[
0\mapsto 10,\qquad
1\mapsto 20,\qquad
2\mapsto 30,\qquad
3\mapsto 50,\qquad
4\mapsto 100
\]
ticks. The main FI-2010 suite evaluates horizons \(0,\dots,4\), while the residual-over-TLOB/BiNCTABL suite evaluates horizons \(\{10,20,50,100\}\) ticks. The training/validation split is chronological within the original training split, with the first 80\% used for training and the last 20\% used for validation. The official testing split is used for test evaluation.
\subsection{Evaluation Metrics and Bootstrap}

Macro-F1 is the primary metric. We also report accuracy, weighted-F1, macro-AUROC, loss, parameter count, trainable parameter count, best validation macro-F1, residual gain \(\gamma\), and elapsed evaluation time. Macro-AUROC is computed from softmax probabilities for single-label tasks and from sigmoid probabilities for multi-label PTB-XL.

For paired comparisons, we use bootstrap resampling with \(B=1000\) samples and bootstrap seed 2026. For a comparison between model \(A\) and model \(B\), the reported bootstrap quantity is
\[
\Delta_{\mathrm{macroF1}}
=
\operatorname{MacroF1}(A)
-
\operatorname{MacroF1}(B).
\]
We report the mean bootstrap delta, the 2.5\% and 97.5\% quantiles, and the empirical probability \(p(\Delta>0)\). Residual models are compared against their corresponding native model, while calibrated models are compared against their uncalibrated base model.

\subsection{Saved Artifacts}

Each run saves the full configuration as JSON, per-seed CSV results, summary CSV files, model checkpoints, compressed validation/test logits, and calibration outputs. Residual logits files include the final logits, frozen native logits, auxiliary logits, scale weights, and residual gain. Calibration files include calibrated test logits and, for search-based methods, selected parameter values and optimization history. FI-2010 also saves parsed file metadata, data split metadata, classification reports, horizon-specific result files, and an all-horizon summary.

\subsection{Computational Resources}

Experiments were run on a workstation with Ubuntu 24.04.4 LTS, Linux kernel 6.8.0-110-generic, an AMD Ryzen Threadripper 7970X CPU with 32 cores and 64 threads, 256 GiB of RAM, and two NVIDIA GeForce RTX 5090 GPUs with 32 GB of VRAM each. The software environment used NVIDIA driver 590.48.01, CUDA Toolkit 12.0 via nvcc 12.0.140, and Python 3.12.3.

\section{Additional Statistical Results}
\label{app:additional_stats}

This section provides additional numerical support for the empirical claims in Section~\ref{sec:experiments}. The appendix tables report two complementary types of evidence. Table~\ref{tab:appendix_best_calibration_compact} summarizes the best post-hoc calibration result for the general temporal classification benchmarks, while Table~\ref{tab:appendix_fi2010_best_calibration_compact} reports the same analysis for FI-2010 across prediction horizons. These tables evaluate whether frozen logits contain decision-level evidence that can be better recombined after training. In contrast, Table~\ref{tab:fi2010_best_adaptation_all_backbones_macro_f1}, Table~\ref{tab:generic_ts_best_adaptation_macro_f1}, Table~\ref{tab:dataset_residual_macro_f1}, and Table~\ref{tab:native_adapters_macro_f1_horizons} evaluate raw residual improvements, measuring whether the residual branch adds useful temporal evidence before any post-hoc calibration is applied. Together, these results support the representation--calibration decomposition used in the main paper.

\paragraph{Calibration results on the general benchmarks.}
Table~\ref{tab:appendix_best_calibration_compact} shows that post-hoc calibration is most useful when the raw model already exposes useful but imperfectly combined evidence. PTB-XL is the clearest example of this effect. For recurrent backbones, calibrating residual models gives substantially larger improvements than calibrating the native model alone. In these cases, the best calibration method is usually a branch-aware separated method, especially EvoCal-S, indicating that preserving native and auxiliary logits separately is useful for decision-level recombination.

For stronger PTB-XL backbones, calibration gains are smaller. ResNet1D remains the strongest PTB-XL backbone, while PatchTransformer and TemporalTransformer show more limited additional headroom after calibration. This supports the interpretation that PTB-XL contains both decision-limited and near-saturated regimes: recurrent models benefit from residual evidence and branch-aware calibration, whereas stronger models leave less decision-level evidence to recover.

HARTH shows a different pattern. Calibration gains are large across several backbones, but they often coincide with large raw residual improvements. This suggests a mixed representation--decision regime: the residual branch changes the available temporal evidence, and calibration further adjusts how this evidence is converted into final predictions. Thus, HARTH should not be interpreted as a purely calibration-driven dataset.

MHEALTH is more variable. The strongest native and residual results are obtained by TemporalTransformer and FCN, where calibration often produces little or no measurable gain over the raw model. In contrast, recurrent backbones benefit more clearly from residual and calibrated variants. The relatively high variability across seeds indicates that this dataset is sensitive to optimization and validation-set fluctuations, so the results should be interpreted as evidence of calibration opportunity rather than as uniformly stable improvements.

UCI-HAR is the closest to a near-saturated regime. Several calibrated models show negligible gains over their raw counterparts, especially for stronger native backbones. The main exceptions occur for weaker recurrent models, where residual evidence and separated calibration can still improve performance. Overall, however, UCI-HAR shows limited remaining headroom compared with FI-2010, PTB-XL, HARTH, and MHEALTH.

\paragraph{Calibration results on FI-2010.}
Table~\ref{tab:appendix_fi2010_best_calibration_compact} shows that calibration gains on FI-2010 are horizon-dependent. At the shortest horizon, calibration produces meaningful additional gains over raw residual models. This indicates that short-horizon limit-order-book prediction is not only representation-sensitive but also decision-sensitive: even after residual evidence is added, the final logits can still be usefully recalibrated.

At intermediate horizons, calibration remains useful but less uniformly so. At longer horizons, the gains become much smaller across most backbone--residual combinations. This pattern supports the main interpretation in Section~\ref{sec:experiments}: as the forecasting horizon increases and the raw residual predictor becomes stronger or more stable, the additional value of post-hoc decision calibration decreases.

\begin{table*}[t]
\centering
\fontsize{4}{4.5}\selectfont
\setlength{\tabcolsep}{2.2pt}
\renewcommand{\arraystretch}{1.12}
\caption{Best post-hoc calibration result by FI-2010 prediction horizon, backbone, and residual variant. Each cell reports test Macro-F1 as mean $\pm$ standard deviation across seeds, in percentage points. The second line reports the winning calibration method and the gain over the corresponding uncalibrated raw model, also in percentage points. Abbreviations: Iso = Isotonic; Hist = Histogram Binning; Vec = Vector Scaling; Mat = Matrix Scaling; Final-RS = final-only random search; Sep-RS = separated random search; EvoCal = evolutionary calibration.}
\label{tab:appendix_fi2010_best_calibration_compact}
\resizebox{0.80\textwidth}{!}{%
\begin{tabular}{llcccc}
\toprule
Horizon & Backbone & Base & Res-MS & Res-Z & Res-RCG \\
\midrule

\multicolumn{6}{l}{\textit{FI-2010, 10 ticks}}\\
 & InceptionTime
 & \makecell[c]{76.53 $\pm$ 1.72\\{\fontsize{4}{4.5}\selectfont EvoCal, $\Delta$ +1.31}}
 & \makecell[c]{83.63 $\pm$ 0.62\\{\fontsize{4}{4.5}\selectfont Sep-RS, $\Delta$ +0.96}}
 & \makecell[c]{82.98 $\pm$ 1.19\\{\fontsize{4}{4.5}\selectfont Final-RS, $\Delta$ +0.80}}
 & \makecell[c]{82.53 $\pm$ 0.63\\{\fontsize{4}{4.5}\selectfont Sep-RS, $\Delta$ +0.52}} \\

 & BiNCTABL
 & \makecell[c]{82.06 $\pm$ 1.42\\{\fontsize{4}{4.5}\selectfont Final-RS, $\Delta$ +1.08}}
 & \makecell[c]{87.12 $\pm$ 1.38\\{\fontsize{4}{4.5}\selectfont Hist, $\Delta$ +0.94}}
 & \makecell[c]{87.16 $\pm$ 1.47\\{\fontsize{4}{4.5}\selectfont Hist, $\Delta$ +0.81}}
 & \makecell[c]{87.16 $\pm$ 1.15\\{\fontsize{4}{4.5}\selectfont Hist, $\Delta$ +0.90}} \\

 & TLOB
 & \makecell[c]{76.12 $\pm$ 0.33\\{\fontsize{4}{4.5}\selectfont Final-RS, $\Delta$ +2.09}}
 & \makecell[c]{82.60 $\pm$ 1.02\\{\fontsize{4}{4.5}\selectfont Final-RS, $\Delta$ +1.32}}
 & \makecell[c]{82.67 $\pm$ 0.06\\{\fontsize{4}{4.5}\selectfont Final-RS, $\Delta$ +1.06}}
 & \makecell[c]{82.88 $\pm$ 0.72\\{\fontsize{4}{4.5}\selectfont Final-RS, $\Delta$ +1.25}} \\

\midrule
\multicolumn{6}{l}{\textit{FI-2010, 20 ticks}}\\
 & InceptionTime
 & \makecell[c]{72.67 $\pm$ 1.83\\{\fontsize{4}{4.5}\selectfont Final-RS, $\Delta$ +0.82}}
 & \makecell[c]{81.39 $\pm$ 0.60\\{\fontsize{4}{4.5}\selectfont Hist, $\Delta$ +0.65}}
 & \makecell[c]{80.80 $\pm$ 0.76\\{\fontsize{4}{4.5}\selectfont Hist, $\Delta$ +0.43}}
 & \makecell[c]{80.07 $\pm$ 0.90\\{\fontsize{4}{4.5}\selectfont EvoCal, $\Delta$ +0.21}} \\

 & BiNCTABL
 & \makecell[c]{74.52 $\pm$ 0.24\\{\fontsize{4}{4.5}\selectfont Final-RS, $\Delta$ +0.90}}
 & \makecell[c]{83.16 $\pm$ 0.31\\{\fontsize{4}{4.5}\selectfont Iso, $\Delta$ +0.54}}
 & \makecell[c]{82.30 $\pm$ 0.66\\{\fontsize{4}{4.5}\selectfont Iso, $\Delta$ +0.61}}
 & \makecell[c]{82.50 $\pm$ 0.21\\{\fontsize{4}{4.5}\selectfont Hist, $\Delta$ +0.69}} \\

 & TLOB
 & \makecell[c]{70.66 $\pm$ 0.06\\{\fontsize{4}{4.5}\selectfont Final-RS, $\Delta$ +1.13}}
 & \makecell[c]{80.20 $\pm$ 0.71\\{\fontsize{4}{4.5}\selectfont Hist, $\Delta$ +0.88}}
 & \makecell[c]{79.87 $\pm$ 0.45\\{\fontsize{4}{4.5}\selectfont Final-RS, $\Delta$ +0.36}}
 & \makecell[c]{80.74 $\pm$ 0.28\\{\fontsize{4}{4.5}\selectfont Final-RS, $\Delta$ +0.57}} \\

\midrule
\multicolumn{6}{l}{\textit{FI-2010, 50 ticks}}\\
 & InceptionTime
 & \makecell[c]{82.41 $\pm$ 1.44\\{\fontsize{4}{4.5}\selectfont EvoCal, $\Delta$ +0.15}}
 & \makecell[c]{88.24 $\pm$ 0.46\\{\fontsize{4}{4.5}\selectfont EvoCal, $\Delta$ +0.03}}
 & \makecell[c]{87.88 $\pm$ 0.42\\{\fontsize{4}{4.5}\selectfont Iso, $\Delta$ +0.03}}
 & \makecell[c]{87.57 $\pm$ 0.53\\{\fontsize{4}{4.5}\selectfont EvoCal, $\Delta$ +0.02}} \\

 & BiNCTABL
 & \makecell[c]{88.67 $\pm$ 0.21\\{\fontsize{4}{4.5}\selectfont Iso, $\Delta$ +0.03}}
 & \makecell[c]{92.17 $\pm$ 0.20\\{\fontsize{4}{4.5}\selectfont Iso, $\Delta$ +0.05}}
 & \makecell[c]{92.01 $\pm$ 0.22\\{\fontsize{4}{4.5}\selectfont Sep-RS, $\Delta$ +0.08}}
 & \makecell[c]{92.14 $\pm$ 0.15\\{\fontsize{4}{4.5}\selectfont Sep-RS, $\Delta$ +0.08}} \\

 & TLOB
 & \makecell[c]{81.82 $\pm$ 0.40\\{\fontsize{4}{4.5}\selectfont Mat, $\Delta$ +0.04}}
 & \makecell[c]{88.44 $\pm$ 0.22\\{\fontsize{4}{4.5}\selectfont Iso, $\Delta$ +0.06}}
 & \makecell[c]{88.08 $\pm$ 0.32\\{\fontsize{4}{4.5}\selectfont Iso, $\Delta$ +0.04}}
 & \makecell[c]{88.02 $\pm$ 0.10\\{\fontsize{4}{4.5}\selectfont Final-RS, $\Delta$ +0.15}} \\

\midrule
\multicolumn{6}{l}{\textit{FI-2010, 100 ticks}}\\
 & InceptionTime
 & \makecell[c]{85.36 $\pm$ 1.02\\{\fontsize{4}{4.5}\selectfont Final-RS, $\Delta$ +0.14}}
 & \makecell[c]{90.05 $\pm$ 0.27\\{\fontsize{4}{4.5}\selectfont Sep-RS, $\Delta$ +0.06}}
 & \makecell[c]{89.66 $\pm$ 0.43\\{\fontsize{4}{4.5}\selectfont Final-RS, $\Delta$ +0.17}}
 & \makecell[c]{89.50 $\pm$ 0.36\\{\fontsize{4}{4.5}\selectfont EvoCal, $\Delta$ +0.13}} \\

 & BiNCTABL
 & \makecell[c]{93.34 $\pm$ 0.10\\{\fontsize{4}{4.5}\selectfont Iso, $\Delta$ +0.17}}
 & \makecell[c]{95.07 $\pm$ 0.20\\{\fontsize{4}{4.5}\selectfont Sep-RS, $\Delta$ +0.12}}
 & \makecell[c]{95.05 $\pm$ 0.07\\{\fontsize{4}{4.5}\selectfont Sep-RS, $\Delta$ +0.10}}
 & \makecell[c]{95.12 $\pm$ 0.23\\{\fontsize{4}{4.5}\selectfont Sep-RS, $\Delta$ +0.12}} \\

 & TLOB
 & \makecell[c]{85.88 $\pm$ 0.17\\{\fontsize{4}{4.5}\selectfont Final-RS, $\Delta$ +0.44}}
 & \makecell[c]{90.44 $\pm$ 0.22\\{\fontsize{4}{4.5}\selectfont Final-RS, $\Delta$ +0.23}}
 & \makecell[c]{90.30 $\pm$ 0.24\\{\fontsize{4}{4.5}\selectfont Final-RS, $\Delta$ +0.20}}
 & \makecell[c]{90.57 $\pm$ 0.20\\{\fontsize{4}{4.5}\selectfont Hist, $\Delta$ +0.13}} \\

\bottomrule
\end{tabular}%
}
\end{table*}

The best calibration method also varies across horizons and backbones. Short horizons often favor search-based or nonparametric calibration, such as separated random search, final-only random search, histogram binning, or EvoCal. Longer horizons show smaller differences among calibration methods. This suggests that no single calibrator dominates uniformly; the benefit comes from using calibration as a diagnostic stage to reveal whether decision-level headroom remains.

\paragraph{Paired residual adaptation on FI-2010.}
Table~\ref{tab:fi2010_best_adaptation_all_backbones_macro_f1} reports paired improvements of the best raw residual adaptation over the corresponding native model on FI-2010. All listed comparisons are positive, and the paired bootstrap confidence intervals exclude zero. The largest improvements occur at the shortest and intermediate horizons, where limit-order-book prediction is more local, noisy, and sensitive to short-term temporal structure.

The gains remain positive at longer horizons but become smaller for the strongest specialized backbone. This supports two conclusions. First, residual multi-scale evidence is consistently useful for FI-2010. Second, the size of the residual opportunity depends on both the prediction horizon and the native backbone: stronger native predictors and more stable horizons leave less unresolved temporal evidence for the residual branch to capture.

\paragraph{Paired residual adaptation on the general benchmarks.}
Table~\ref{tab:generic_ts_best_adaptation_macro_f1} reports paired raw residual gains for PTB-XL, HARTH, MHEALTH, and UCI-HAR. The strongest improvements occur for recurrent backbones, especially in HARTH, MHEALTH, and PTB-XL. These results support the claim that residual multi-scale evidence is most useful when the native backbone is representation-limited.

For stronger convolutional, residual, and Transformer-based backbones, raw residual gains are smaller and sometimes negligible. UCI-HAR follows the same pattern: the best residual gains are modest, while several stronger models change very little. This supports the near-saturation interpretation for UCI-HAR and for stronger backbones on the general benchmarks.

\paragraph{Raw residual variants on the general benchmarks.}
Table~\ref{tab:dataset_residual_macro_f1} compares Native, Residual-MS, Residual-ZOnly, and Residual-RCG directly. The table shows that the best trust-allocation rule is dataset- and backbone-dependent. On PTB-XL, Residual-ZOnly is the best residual variant for most backbones, suggesting that learned scale representations often already encode enough information for trust allocation without requiring explicit reliability descriptors.

On HARTH and MHEALTH, Residual-RCG is more useful for unstable recurrent models. This suggests that explicit reliability descriptors help when temporal instability is not already captured by the learned scale representations. However, RCG is not uniformly best: some recurrent and stronger-backbone settings favor ZOnly or MS, and several strong native models show negligible differences among the residual variants. This reinforces the conclusion that explicit reliability descriptors are useful in specific regimes rather than universally superior.

\begin{table*}[t]
\centering
\fontsize{4}{4.5}\selectfont
\setlength{\tabcolsep}{2.2pt}
\renewcommand{\arraystretch}{1.12}
\caption{Best post-hoc calibration result by dataset, backbone, and residual variant. Each cell reports test Macro-F1 as mean $\pm$ standard deviation across seeds, in percentage points. The second line reports the winning calibration method and the gain over the corresponding uncalibrated raw model, also in percentage points. Abbreviations: AdaT = Adaptive Temperature; Iso = Isotonic; Hist = Histogram Binning; Vec = Vector Scaling; Mat = Matrix Scaling; Final-RS = final-only random search; Sep-RS = separated random search; EvoCal-F = final-only EvoCal; EvoCal-S = separated EvoCal.}
\label{tab:appendix_best_calibration_compact}
\resizebox{0.80\textwidth}{!}{%
\begin{tabular}{llcccc}
\toprule
Dataset & Backbone & Base & Res-MS & Res-Z & Res-RCG \\
\midrule
\multicolumn{6}{l}{\textit{PTB-XL}}\\
 & FCN & \makecell[c]{71.10 $\pm$ 0.42\\{\fontsize{4}{4.5}\selectfont EvoCal-F, $\Delta$ +0.90}} & \makecell[c]{71.62 $\pm$ 0.43\\{\fontsize{4}{4.5}\selectfont EvoCal-S, $\Delta$ +1.29}} & \makecell[c]{71.57 $\pm$ 0.48\\{\fontsize{4}{4.5}\selectfont EvoCal-S, $\Delta$ +0.90}} & \makecell[c]{71.55 $\pm$ 0.28\\{\fontsize{4}{4.5}\selectfont EvoCal-F, $\Delta$ +1.19}} \\
 & GRU & \makecell[c]{59.44 $\pm$ 2.54\\{\fontsize{4}{4.5}\selectfont EvoCal-F, $\Delta$ +0.86}} & \makecell[c]{63.79 $\pm$ 1.17\\{\fontsize{4}{4.5}\selectfont EvoCal-S, $\Delta$ +3.01}} & \makecell[c]{64.05 $\pm$ 0.63\\{\fontsize{4}{4.5}\selectfont EvoCal-S, $\Delta$ +2.50}} & \makecell[c]{64.30 $\pm$ 1.03\\{\fontsize{4}{4.5}\selectfont EvoCal-S, $\Delta$ +2.85}} \\
 & LSTM & \makecell[c]{55.78 $\pm$ 1.00\\{\fontsize{4}{4.5}\selectfont EvoCal-F, $\Delta$ +0.59}} & \makecell[c]{62.08 $\pm$ 0.76\\{\fontsize{4}{4.5}\selectfont EvoCal-S, $\Delta$ +4.13}} & \makecell[c]{63.43 $\pm$ 1.15\\{\fontsize{4}{4.5}\selectfont EvoCal-S, $\Delta$ +4.54}} & \makecell[c]{62.84 $\pm$ 0.72\\{\fontsize{4}{4.5}\selectfont EvoCal-S, $\Delta$ +3.99}} \\
 & PatchTransf. & \makecell[c]{69.70 $\pm$ 0.86\\{\fontsize{4}{4.5}\selectfont EvoCal-F, $\Delta$ +0.70}} & \makecell[c]{69.84 $\pm$ 0.75\\{\fontsize{4}{4.5}\selectfont EvoCal-F, $\Delta$ +0.85}} & \makecell[c]{69.70 $\pm$ 0.68\\{\fontsize{4}{4.5}\selectfont EvoCal-F, $\Delta$ +0.62}} & \makecell[c]{69.74 $\pm$ 0.79\\{\fontsize{4}{4.5}\selectfont EvoCal-S, $\Delta$ +0.73}} \\
 & ResNet1D & \makecell[c]{74.24 $\pm$ 0.29\\{\fontsize{4}{4.5}\selectfont EvoCal-F, $\Delta$ +0.71}} & \makecell[c]{74.57 $\pm$ 0.58\\{\fontsize{4}{4.5}\selectfont EvoCal-S, $\Delta$ +0.67}} & \makecell[c]{74.59 $\pm$ 0.23\\{\fontsize{4}{4.5}\selectfont EvoCal-F, $\Delta$ +0.66}} & \makecell[c]{74.46 $\pm$ 0.47\\{\fontsize{4}{4.5}\selectfont Final-RS, $\Delta$ +0.54}} \\
 & TempTransf. & \makecell[c]{67.17 $\pm$ 0.39\\{\fontsize{4}{4.5}\selectfont Final-RS, $\Delta$ +0.81}} & \makecell[c]{67.05 $\pm$ 0.65\\{\fontsize{4}{4.5}\selectfont EvoCal-S, $\Delta$ +0.74}} & \makecell[c]{67.06 $\pm$ 0.63\\{\fontsize{4}{4.5}\selectfont EvoCal-F, $\Delta$ +0.72}} & \makecell[c]{67.10 $\pm$ 0.76\\{\fontsize{4}{4.5}\selectfont EvoCal-S, $\Delta$ +0.74}} \\

\midrule
\multicolumn{6}{l}{\textit{HARTH}}\\
 & FCN & \makecell[c]{73.23 $\pm$ 1.36\\{\fontsize{4}{4.5}\selectfont Iso, $\Delta$ +6.62}} & \makecell[c]{73.97 $\pm$ 0.60\\{\fontsize{4}{4.5}\selectfont Iso, $\Delta$ +7.71}} & \makecell[c]{73.64 $\pm$ 0.63\\{\fontsize{4}{4.5}\selectfont Iso, $\Delta$ +7.23}} & \makecell[c]{73.28 $\pm$ 1.15\\{\fontsize{4}{4.5}\selectfont Iso, $\Delta$ +6.83}} \\
 & GRU & \makecell[c]{51.27 $\pm$ 5.97\\{\fontsize{4}{4.5}\selectfont Vec, $\Delta$ +5.89}} & \makecell[c]{65.12 $\pm$ 10.30\\{\fontsize{4}{4.5}\selectfont Mat, $\Delta$ +17.40}} & \makecell[c]{56.44 $\pm$ 6.37\\{\fontsize{4}{4.5}\selectfont EvoCal-S, $\Delta$ +7.57}} & \makecell[c]{69.71 $\pm$ 6.18\\{\fontsize{4}{4.5}\selectfont Vec, $\Delta$ +12.13}} \\
 & LSTM & \makecell[c]{49.92 $\pm$ 1.31\\{\fontsize{4}{4.5}\selectfont Mat, $\Delta$ +7.60}} & \makecell[c]{65.29 $\pm$ 10.71\\{\fontsize{4}{4.5}\selectfont Mat, $\Delta$ +13.11}} & \makecell[c]{67.90 $\pm$ 3.37\\{\fontsize{4}{4.5}\selectfont Vec, $\Delta$ +9.99}} & \makecell[c]{68.88 $\pm$ 6.06\\{\fontsize{4}{4.5}\selectfont Vec, $\Delta$ +12.23}} \\
 & PatchTransf. & \makecell[c]{79.33 $\pm$ 3.37\\{\fontsize{4}{4.5}\selectfont Vec, $\Delta$ +13.28}} & \makecell[c]{80.43 $\pm$ 2.22\\{\fontsize{4}{4.5}\selectfont Vec, $\Delta$ +14.50}} & \makecell[c]{79.40 $\pm$ 3.49\\{\fontsize{4}{4.5}\selectfont Vec, $\Delta$ +13.47}} & \makecell[c]{79.20 $\pm$ 3.49\\{\fontsize{4}{4.5}\selectfont Vec, $\Delta$ +13.67}} \\
 & ResNet1D & \makecell[c]{76.34 $\pm$ 9.24\\{\fontsize{4}{4.5}\selectfont Hist, $\Delta$ +11.92}} & \makecell[c]{76.54 $\pm$ 6.28\\{\fontsize{4}{4.5}\selectfont Hist, $\Delta$ +12.14}} & \makecell[c]{74.05 $\pm$ 9.36\\{\fontsize{4}{4.5}\selectfont Hist, $\Delta$ +8.08}} & \makecell[c]{74.42 $\pm$ 10.00\\{\fontsize{4}{4.5}\selectfont Hist, $\Delta$ +10.03}} \\
 & TempTransf. & \makecell[c]{75.04 $\pm$ 3.20\\{\fontsize{4}{4.5}\selectfont Iso, $\Delta$ +12.03}} & \makecell[c]{74.98 $\pm$ 4.96\\{\fontsize{4}{4.5}\selectfont Vec, $\Delta$ +11.89}} & \makecell[c]{75.48 $\pm$ 2.93\\{\fontsize{4}{4.5}\selectfont Iso, $\Delta$ +12.27}} & \makecell[c]{75.55 $\pm$ 2.91\\{\fontsize{4}{4.5}\selectfont Iso, $\Delta$ +12.39}} \\

\midrule
\multicolumn{6}{l}{\textit{MHEALTH}}\\
 & FCN & \makecell[c]{89.58 $\pm$ 4.40\\{\fontsize{4}{4.5}\selectfont AdaT, $\Delta$ +0.00}} & \makecell[c]{90.43 $\pm$ 3.28\\{\fontsize{4}{4.5}\selectfont AdaT, $\Delta$ +0.00}} & \makecell[c]{90.14 $\pm$ 3.04\\{\fontsize{4}{4.5}\selectfont AdaT, $\Delta$ +0.00}} & \makecell[c]{90.37 $\pm$ 3.50\\{\fontsize{4}{4.5}\selectfont AdaT, $\Delta$ +0.00}} \\
 & GRU & \makecell[c]{54.99 $\pm$ 15.30\\{\fontsize{4}{4.5}\selectfont Vec, $\Delta$ +3.37}} & \makecell[c]{57.46 $\pm$ 18.44\\{\fontsize{4}{4.5}\selectfont Vec, $\Delta$ +1.66}} & \makecell[c]{62.79 $\pm$ 18.91\\{\fontsize{4}{4.5}\selectfont EvoCal-S, $\Delta$ +6.86}} & \makecell[c]{70.23 $\pm$ 10.35\\{\fontsize{4}{4.5}\selectfont EvoCal-S, $\Delta$ +6.41}} \\
 & LSTM & \makecell[c]{62.50 $\pm$ 9.16\\{\fontsize{4}{4.5}\selectfont Vec, $\Delta$ +0.96}} & \makecell[c]{70.25 $\pm$ 3.32\\{\fontsize{4}{4.5}\selectfont EvoCal-S, $\Delta$ +2.70}} & \makecell[c]{66.02 $\pm$ 7.40\\{\fontsize{4}{4.5}\selectfont Iso, $\Delta$ +0.74}} & \makecell[c]{69.45 $\pm$ 7.20\\{\fontsize{4}{4.5}\selectfont Sep-RS, $\Delta$ +1.30}} \\
 & PatchTransf. & \makecell[c]{88.95 $\pm$ 1.26\\{\fontsize{4}{4.5}\selectfont AdaT, $\Delta$ +0.00}} & \makecell[c]{89.01 $\pm$ 1.33\\{\fontsize{4}{4.5}\selectfont AdaT, $\Delta$ +0.00}} & \makecell[c]{88.95 $\pm$ 1.26\\{\fontsize{4}{4.5}\selectfont AdaT, $\Delta$ +0.00}} & \makecell[c]{89.84 $\pm$ 1.84\\{\fontsize{4}{4.5}\selectfont Sep-RS, $\Delta$ +0.77}} \\
 & ResNet1D & \makecell[c]{83.69 $\pm$ 5.14\\{\fontsize{4}{4.5}\selectfont AdaT, $\Delta$ +0.00}} & \makecell[c]{85.70 $\pm$ 4.31\\{\fontsize{4}{4.5}\selectfont Sep-RS, $\Delta$ +1.66}} & \makecell[c]{84.26 $\pm$ 4.51\\{\fontsize{4}{4.5}\selectfont AdaT, $\Delta$ +0.00}} & \makecell[c]{84.12 $\pm$ 4.75\\{\fontsize{4}{4.5}\selectfont EvoCal-S, $\Delta$ +0.04}} \\
 & TempTransf. & \makecell[c]{94.57 $\pm$ 2.76\\{\fontsize{4}{4.5}\selectfont AdaT, $\Delta$ +0.00}} & \makecell[c]{94.38 $\pm$ 2.57\\{\fontsize{4}{4.5}\selectfont AdaT, $\Delta$ +0.00}} & \makecell[c]{94.50 $\pm$ 2.59\\{\fontsize{4}{4.5}\selectfont AdaT, $\Delta$ +0.00}} & \makecell[c]{94.57 $\pm$ 2.76\\{\fontsize{4}{4.5}\selectfont AdaT, $\Delta$ +0.00}} \\

\midrule
\multicolumn{6}{l}{\textit{UCI-HAR}}\\
 & FCN & \makecell[c]{92.33 $\pm$ 0.39\\{\fontsize{4}{4.5}\selectfont AdaT, $\Delta$ +0.00}} & \makecell[c]{92.33 $\pm$ 0.44\\{\fontsize{4}{4.5}\selectfont AdaT, $\Delta$ +0.00}} & \makecell[c]{91.99 $\pm$ 0.28\\{\fontsize{4}{4.5}\selectfont AdaT, $\Delta$ +0.00}} & \makecell[c]{92.39 $\pm$ 0.35\\{\fontsize{4}{4.5}\selectfont AdaT, $\Delta$ +0.00}} \\
 & GRU & \makecell[c]{87.66 $\pm$ 1.47\\{\fontsize{4}{4.5}\selectfont Final-RS, $\Delta$ +0.04}} & \makecell[c]{88.32 $\pm$ 1.46\\{\fontsize{4}{4.5}\selectfont AdaT, $\Delta$ +0.00}} & \makecell[c]{88.70 $\pm$ 0.22\\{\fontsize{4}{4.5}\selectfont Sep-RS, $\Delta$ +0.45}} & \makecell[c]{88.38 $\pm$ 0.88\\{\fontsize{4}{4.5}\selectfont AdaT, $\Delta$ +0.00}} \\
 & LSTM & \makecell[c]{84.06 $\pm$ 3.11\\{\fontsize{4}{4.5}\selectfont Final-RS, $\Delta$ +0.09}} & \makecell[c]{85.65 $\pm$ 3.16\\{\fontsize{4}{4.5}\selectfont EvoCal-S, $\Delta$ +0.72}} & \makecell[c]{88.45 $\pm$ 1.31\\{\fontsize{4}{4.5}\selectfont Sep-RS, $\Delta$ +2.67}} & \makecell[c]{89.00 $\pm$ 3.19\\{\fontsize{4}{4.5}\selectfont Sep-RS, $\Delta$ +4.27}} \\
 & PatchTransf. & \makecell[c]{91.65 $\pm$ 0.44\\{\fontsize{4}{4.5}\selectfont AdaT, $\Delta$ +0.00}} & \makecell[c]{91.63 $\pm$ 0.59\\{\fontsize{4}{4.5}\selectfont Final-RS, $\Delta$ +0.05}} & \makecell[c]{91.62 $\pm$ 0.42\\{\fontsize{4}{4.5}\selectfont AdaT, $\Delta$ +0.00}} & \makecell[c]{91.82 $\pm$ 0.30\\{\fontsize{4}{4.5}\selectfont Final-RS, $\Delta$ +0.23}} \\
 & ResNet1D & \makecell[c]{92.97 $\pm$ 1.58\\{\fontsize{4}{4.5}\selectfont AdaT, $\Delta$ +0.00}} & \makecell[c]{93.15 $\pm$ 1.60\\{\fontsize{4}{4.5}\selectfont AdaT, $\Delta$ +0.00}} & \makecell[c]{92.84 $\pm$ 1.40\\{\fontsize{4}{4.5}\selectfont AdaT, $\Delta$ +0.00}} & \makecell[c]{93.06 $\pm$ 1.46\\{\fontsize{4}{4.5}\selectfont AdaT, $\Delta$ +0.00}} \\
 & TempTransf. & \makecell[c]{89.83 $\pm$ 0.91\\{\fontsize{4}{4.5}\selectfont Iso, $\Delta$ +0.26}} & \makecell[c]{89.81 $\pm$ 0.91\\{\fontsize{4}{4.5}\selectfont Sep-RS, $\Delta$ +0.36}} & \makecell[c]{89.92 $\pm$ 0.80\\{\fontsize{4}{4.5}\selectfont Sep-RS, $\Delta$ +0.46}} & \makecell[c]{89.92 $\pm$ 1.01\\{\fontsize{4}{4.5}\selectfont Sep-RS, $\Delta$ +0.44}} \\
\bottomrule
\end{tabular}%
}
\end{table*}

\paragraph{Raw residual variants on FI-2010.}
Table~\ref{tab:native_adapters_macro_f1_horizons} reports raw FI-2010 results by horizon. The main pattern is that Residual-MS is a strong default on this benchmark. It is the best variant for all InceptionTime horizons and for several backbone--horizon combinations, indicating that much of the gain comes from adding multi-scale residual evidence itself rather than from learning a more complex trust-allocation rule.

There are, however, cases where adaptive trust helps. RCG and ZOnly are strongest for some TLOB and BiNCTABL settings, especially when the useful temporal scale appears to vary more strongly across examples. These cases suggest that adaptive or reliability-aware trust can help, but the effect is not uniform across horizons or backbones.

Across FI-2010, the magnitude of the residual gain is largest at short and intermediate horizons and remains positive at longer horizons. This confirms that FI-2010 is a residual-sensitive benchmark, especially when the prediction problem is local, noisy, and difficult.

\paragraph{Summary of appendix evidence.}
The appendix results support the four main claims of the paper. First, residual evidence is useful when the native model is representation-limited, as shown by consistent FI-2010 gains and large recurrent-backbone gains on HARTH, MHEALTH, and PTB-XL. Second, learned trust allocation is useful but not universally superior: ZOnly is often sufficient on PTB-XL, RCG helps in unstable recurrent settings, and MS is a strong default on FI-2010. Third, calibration is most useful when the logits contain underused decision-level evidence, especially in PTB-XL recurrent models and short-horizon FI-2010. Fourth, the datasets separate into regimes: FI-2010 is residual-sensitive, PTB-XL is partly calibration-sensitive, HARTH and MHEALTH are backbone-dependent and often unstable, and UCI-HAR is closest to saturation.
\begin{table*}[t]
\centering
\caption{
Best paired adaptation against the corresponding raw/native model on FI-2010.
Results report mean macro-F1 over three seeds. $\Delta$ denotes the paired
improvement over the corresponding unadapted model. Confidence intervals are
95\% paired bootstrap CIs over seeds. Green cells indicate that the bootstrap
CI for $\Delta$ excludes zero.
}
\label{tab:fi2010_best_adaptation_all_backbones_macro_f1}
\resizebox{\textwidth}{!}{
\begin{tabular}{ll l c c c c}
\toprule
Horizon & Backbone & Best adaptation & Raw/Native & Adapted & $\Delta$ macro-F1 & 95\% CI \\
\midrule
10  & TLOB          & RCG   & 0.7403 & 0.8163 & \bsig{+0.0760} & \bsig{[0.0647, 0.0821]} \\
10  & BiNCTABL      & ZOnly & 0.8098 & 0.8636 & \bsig{+0.0538} & \bsig{[0.0439, 0.0588]} \\
10  & InceptionTime & MS    & 0.7521 & 0.8267 & \bsig{+0.0746} & \bsig{[0.0697, 0.0801]} \\
\midrule
20  & TLOB          & RCG   & 0.6953 & 0.8017 & \bsig{+0.1065} & \bsig{[0.0986, 0.1107]} \\
20  & BiNCTABL      & MS    & 0.7363 & 0.8261 & \bsig{+0.0899} & \bsig{[0.0878, 0.0918]} \\
20  & InceptionTime & MS    & 0.7185 & 0.8074 & \bsig{+0.0889} & \bsig{[0.0720, 0.0984]} \\
\midrule
50  & TLOB          & MS    & 0.8178 & 0.8837 & \bsig{+0.0659} & \bsig{[0.0604, 0.0705]} \\
50  & BiNCTABL      & MS    & 0.8864 & 0.9212 & \bsig{+0.0348} & \bsig{[0.0343, 0.0353]} \\
50  & InceptionTime & MS    & 0.8226 & 0.8821 & \bsig{+0.0595} & \bsig{[0.0482, 0.0696]} \\
\midrule
100 & TLOB          & RCG   & 0.8544 & 0.9044 & \bsig{+0.0500} & \bsig{[0.0473, 0.0528]} \\
100 & BiNCTABL      & RCG   & 0.9318 & 0.9501 & \bsig{+0.0183} & \bsig{[0.0163, 0.0205]} \\
100 & InceptionTime & MS    & 0.8522 & 0.8999 & \bsig{+0.0477} & \bsig{[0.0382, 0.0550]} \\
\bottomrule
\end{tabular}
}
\vspace{0.5em}

\end{table*}

\begin{table*}[t]
\centering
\caption{
Best residual adaptation against the corresponding raw baseline on the general
time-series benchmarks. Results report mean macro-F1 over three seeds.
$\Delta$ denotes the paired improvement over the raw model. Confidence
intervals are 95\% paired bootstrap CIs over seeds. Green cells indicate that
the bootstrap CI for $\Delta$ excludes zero.
}
\label{tab:generic_ts_best_adaptation_macro_f1}
\resizebox{\textwidth}{!}{
\begin{tabular}{ll l c c c c}
\toprule
Dataset & Backbone & Best adaptation & Raw & Adapted & $\Delta$ macro-F1 & 95\% CI \\
\midrule
HARTH & FCN                 & RCG   & 0.6661 & 0.6646 & -0.0015 & [-0.0055, 0.0034] \\
HARTH & GRU                 & RCG   & 0.4537 & 0.5758 & \bsig{+0.1221} & \bsig{[0.1082, 0.1322]} \\
HARTH & LSTM                & ZOnly & 0.4231 & 0.5791 & \bsig{+0.1560} & \bsig{[0.1446, 0.1663]} \\
HARTH & PatchTransformer    & MS    & 0.6606 & 0.6593 & -0.0013 & [-0.0093, 0.0046] \\
HARTH & ResNet1D            & ZOnly & 0.6441 & 0.6598 & +0.0156 & [-0.0056, 0.0576] \\
HARTH & TemporalTransformer & ZOnly & 0.6301 & 0.6321 & \bsig{+0.0020} & \bsig{[0.0005, 0.0044]} \\
\midrule
MHEALTH & FCN                 & MS    & 0.8958 & 0.9043 & +0.0086 & [-0.0002, 0.0223] \\
MHEALTH & GRU                 & RCG   & 0.5162 & 0.6383 & \bsig{+0.1220} & \bsig{[0.0735, 0.1511]} \\
MHEALTH & LSTM                & RCG   & 0.6154 & 0.6815 & \bsig{+0.0661} & \bsig{[0.0168, 0.1116]} \\
MHEALTH & PatchTransformer    & RCG   & 0.8895 & 0.8907 & +0.0012 & [0.0000, 0.0036] \\
MHEALTH & ResNet1D            & ZOnly & 0.8369 & 0.8426 & +0.0056 & [0.0000, 0.0136] \\
MHEALTH & TemporalTransformer & RCG   & 0.9457 & 0.9457 & 0.0000 & [0.0000, 0.0000] \\
\midrule
PTB-XL & FCN                 & ZOnly & 0.7020 & 0.7067 & +0.0047 & [-0.0007, 0.0111] \\
PTB-XL & GRU                 & ZOnly & 0.5858 & 0.6155 & \bsig{+0.0297} & \bsig{[0.0223, 0.0383]} \\
PTB-XL & LSTM                & ZOnly & 0.5519 & 0.5889 & \bsig{+0.0369} & \bsig{[0.0347, 0.0407]} \\
PTB-XL & PatchTransformer    & ZOnly & 0.6900 & 0.6908 & +0.0008 & [-0.0001, 0.0025] \\
PTB-XL & ResNet1D            & ZOnly & 0.7353 & 0.7393 & +0.0040 & [-0.0050, 0.0087] \\
PTB-XL & TemporalTransformer & RCG   & 0.6637 & 0.6636 & -0.0001 & [-0.0003, 0.0002] \\
\midrule
UCI-HAR & FCN                 & RCG   & 0.9233 & 0.9239 & +0.0006 & [-0.0021, 0.0034] \\
UCI-HAR & GRU                 & RCG   & 0.8763 & 0.8838 & \bsig{+0.0075} & \bsig{[0.0024, 0.0120]} \\
UCI-HAR & LSTM                & ZOnly & 0.8397 & 0.8578 & \bsig{+0.0181} & \bsig{[0.0010, 0.0422]} \\
UCI-HAR & PatchTransformer    & ZOnly & 0.9165 & 0.9162 & -0.0002 & [-0.0011, 0.0004] \\
UCI-HAR & ResNet1D            & MS    & 0.9297 & 0.9315 & +0.0019 & [-0.0012, 0.0067] \\
UCI-HAR & TemporalTransformer & RCG   & 0.8957 & 0.8948 & \bsig{-0.0009} & \bsig{[-0.0017, -0.0000]} \\
\bottomrule
\end{tabular}
}
\vspace{0.5em}

\end{table*}

This section provides the full numerical support for the residual-gain and dataset-regime analysis discussed in Section~\ref{sec:experiments}. Table~\ref{tab:dataset_residual_macro_f1} reports the corresponding detailed results.

\begin{table*}[t]
\centering
% \scriptsize
\setlength{\tabcolsep}{3pt}
\renewcommand{\arraystretch}{0.82}
\caption{Macro-F1 (\%) comparison between native models and residual variants. Values are mean $\pm$ standard deviation over three seeds. Native results are shown in blue, residual variants in red. The best result per row is bolded and the second best is underlined. $\Delta_{\mathrm{best}}$ denotes the difference between the best residual variant and the native model, in percentage points.}
\label{tab:dataset_residual_macro_f1}

\newcommand{\native}[1]{\textcolor{blue}{#1}}
\newcommand{\calib}[1]{\textcolor{red}{#1}}
\newcommand{\best}[1]{\textbf{#1}}
\newcommand{\second}[1]{\uline{#1}}

% ============================================================
% PTB-XL
% ============================================================
\begin{subtable}{\textwidth}
\centering
\caption{PTB-XL}
\resizebox{1\textwidth}{!}{
\begin{tabular}{lccccc}
\toprule
Model & Native & MS & ZOnly & RCG & $\Delta_{\mathrm{best}}$ \\
\midrule
FCN &
\native{70.20 $\pm$ 0.40} &
\calib{70.33 $\pm$ 0.51} &
\best{\calib{70.67 $\pm$ 0.36}} &
\second{\calib{70.35 $\pm$ 0.33}} &
+0.47 \\
GRU &
\native{58.58 $\pm$ 1.84} &
\calib{60.78 $\pm$ 1.43} &
\best{\calib{61.55 $\pm$ 1.10}} &
\second{\calib{61.45 $\pm$ 0.93}} &
+2.97 \\
LSTM &
\native{55.19 $\pm$ 0.56} &
\calib{57.94 $\pm$ 0.43} &
\best{\calib{58.89 $\pm$ 0.74}} &
\second{\calib{58.85 $\pm$ 0.18}} &
+3.69 \\
PatchTransformer &
\native{69.00 $\pm$ 1.03} &
\calib{68.99 $\pm$ 0.92} &
\best{\calib{69.08 $\pm$ 0.89}} &
\second{\calib{69.01 $\pm$ 0.91}} &
+0.08 \\
ResNet1D &
\native{73.53 $\pm$ 0.50} &
\calib{73.90 $\pm$ 0.45} &
\best{\calib{73.93 $\pm$ 0.46}} &
\second{\calib{73.92 $\pm$ 0.52}} &
+0.40 \\
TemporalTransformer &
\best{\native{66.37 $\pm$ 0.63}} &
\calib{66.31 $\pm$ 0.65} &
\calib{66.34 $\pm$ 0.67} &
\second{\calib{66.36 $\pm$ 0.66}} &
-0.01 \\
\bottomrule
\end{tabular}
}
\end{subtable}

\vspace{0.8em}

% ============================================================
% HARTH
% ============================================================
\begin{subtable}{\textwidth}
\centering
\caption{HARTH}
\resizebox{1\textwidth}{!}{
\begin{tabular}{lccccc}
\toprule
Model & Native & MS & ZOnly & RCG & $\Delta_{\mathrm{best}}$ \\
\midrule
FCN &
\best{\native{66.61 $\pm$ 2.85}} &
\calib{66.26 $\pm$ 1.43} &
\calib{66.41 $\pm$ 2.65} &
\second{\calib{66.46 $\pm$ 2.89}} &
-0.15 \\
GRU &
\native{45.37 $\pm$ 0.18} &
\calib{47.72 $\pm$ 2.91} &
\second{\calib{48.87 $\pm$ 6.98}} &
\best{\calib{57.58 $\pm$ 1.25}} &
+12.21 \\
LSTM &
\native{42.31 $\pm$ 1.54} &
\calib{52.18 $\pm$ 8.05} &
\best{\calib{57.91 $\pm$ 0.78}} &
\second{\calib{56.65 $\pm$ 2.47}} &
+15.60 \\
PatchTransformer &
\best{\native{66.06 $\pm$ 1.80}} &
\second{\calib{65.93 $\pm$ 2.48}} &
\calib{65.93 $\pm$ 2.09} &
\calib{65.53 $\pm$ 3.07} &
-0.13 \\
ResNet1D &
\second{\native{64.41 $\pm$ 3.65}} &
\calib{64.40 $\pm$ 3.52} &
\best{\calib{65.98 $\pm$ 7.25}} &
\calib{64.39 $\pm$ 3.79} &
+1.56 \\
TemporalTransformer &
\native{63.01 $\pm$ 2.47} &
\calib{63.09 $\pm$ 2.37} &
\best{\calib{63.21 $\pm$ 2.45}} &
\second{\calib{63.16 $\pm$ 2.44}} &
+0.20 \\
\bottomrule
\end{tabular}
}
\end{subtable}

\vspace{0.8em}

% ============================================================
% MHEALTH
% ============================================================
\begin{subtable}{\textwidth}
\centering
\caption{MHEALTH}
\resizebox{1\textwidth}{!}{
\begin{tabular}{lccccc}
\toprule
Model & Native & MS & ZOnly & RCG & $\Delta_{\mathrm{best}}$ \\
\midrule
FCN &
\native{89.58 $\pm$ 4.40} &
\best{\calib{90.43 $\pm$ 3.28}} &
\calib{90.14 $\pm$ 3.04} &
\second{\calib{90.37 $\pm$ 3.50}} &
+0.86 \\
GRU &
\native{51.62 $\pm$ 11.37} &
\calib{55.80 $\pm$ 17.90} &
\second{\calib{55.93 $\pm$ 17.02}} &
\best{\calib{63.83 $\pm$ 15.46}} &
+12.20 \\
LSTM &
\native{61.54 $\pm$ 10.95} &
\second{\calib{67.55 $\pm$ 9.84}} &
\calib{65.28 $\pm$ 10.52} &
\best{\calib{68.15 $\pm$ 8.06}} &
+6.61 \\
PatchTransformer &
\native{88.95 $\pm$ 1.26} &
\second{\calib{89.01 $\pm$ 1.33}} &
\calib{88.95 $\pm$ 1.26} &
\best{\calib{89.07 $\pm$ 1.39}} &
+0.12 \\
ResNet1D &
\native{83.69 $\pm$ 5.14} &
\calib{84.04 $\pm$ 4.05} &
\best{\calib{84.26 $\pm$ 4.51}} &
\second{\calib{84.08 $\pm$ 4.15}} &
+0.56 \\
TemporalTransformer &
\best{\native{94.57 $\pm$ 2.76}} &
\calib{94.38 $\pm$ 2.57} &
\second{\calib{94.50 $\pm$ 2.59}} &
\best{\calib{94.57 $\pm$ 2.76}} &
+0.00 \\
\bottomrule
\end{tabular}
}
\end{subtable}

\vspace{0.8em}

% ============================================================
% UCI_HAR
% ============================================================
\begin{subtable}{\textwidth}
\centering
\small
\caption{UCI\_HAR}
\resizebox{1\textwidth}{!}{
\begin{tabular}{lccccc}
\toprule
Model & Native & MS & ZOnly & RCG & $\Delta_{\mathrm{best}}$ \\
\midrule
FCN &
\native{92.33 $\pm$ 0.39} &
\second{\calib{92.33 $\pm$ 0.44}} &
\calib{91.99 $\pm$ 0.28} &
\best{\calib{92.39 $\pm$ 0.35}} &
+0.06 \\
GRU &
\native{87.63 $\pm$ 1.26} &
\second{\calib{88.32 $\pm$ 1.46}} &
\calib{88.25 $\pm$ 1.49} &
\best{\calib{88.38 $\pm$ 0.88}} &
+0.75 \\
LSTM &
\native{83.97 $\pm$ 2.87} &
\second{\calib{84.93 $\pm$ 1.95}} &
\best{\calib{85.78 $\pm$ 1.40}} &
\calib{84.72 $\pm$ 2.12} &
+1.81 \\
PatchTransformer &
\best{\native{91.65 $\pm$ 0.44}} &
\calib{91.58 $\pm$ 0.40} &
\second{\calib{91.62 $\pm$ 0.42}} &
\calib{91.59 $\pm$ 0.42} &
-0.02 \\
ResNet1D &
\native{92.97 $\pm$ 1.58} &
\best{\calib{93.15 $\pm$ 1.60}} &
\calib{92.84 $\pm$ 1.40} &
\second{\calib{93.06 $\pm$ 1.46}} &
+0.19 \\
TemporalTransformer &
\best{\native{89.57 $\pm$ 1.13}} &
\calib{89.45 $\pm$ 1.07} &
\calib{89.46 $\pm$ 1.09} &
\second{\calib{89.48 $\pm$ 1.13}} &
-0.09 \\
\bottomrule
\end{tabular}
}
\end{subtable}

\end{table*}

% Requires:
% \usepackage{booktabs}
% \usepackage{subcaption}
% \usepackage{graphicx}
% \usepackage{xcolor}
% \usepackage[normalem]{ulem}

\newcommand{\native}[1]{\textcolor{blue}{#1}}
\newcommand{\calib}[1]{\textcolor{red}{#1}}
\newcommand{\best}[1]{\textbf{#1}}
\newcommand{\second}[1]{\uline{#1}}

\begin{table*}[t]
\centering
\caption{Macro-F1 of native models and adapters across prediction horizons. Values are reported as mean $\pm$ standard deviation in percentage points.}
\label{tab:native_adapters_macro_f1_horizons}

\large
\renewcommand{\arraystretch}{1.35}
\setlength{\tabcolsep}{10pt}

% ============================================================
% h = 10
% ============================================================
\begin{subtable}{\textwidth}
\centering
\caption{$h=10$}
\resizebox{\textwidth}{!}{
\begin{tabular}{lccccc}
\toprule
Model & Native & MS & ZOnly & RCG & $\Delta_{\mathrm{best}}$ \\
\midrule
BiNCTABL &
\native{80.98 $\pm$ 0.76} &
\calib{86.18 $\pm$ 1.28} &
\best{\calib{86.36 $\pm$ 1.62}} &
\second{\calib{86.27 $\pm$ 1.23}} &
+5.38 \\
TLOB &
\native{74.03 $\pm$ 0.24} &
\calib{81.28 $\pm$ 1.02} &
\second{\calib{81.61 $\pm$ 0.08}} &
\best{\calib{81.63 $\pm$ 0.87}} &
+7.60 \\
InceptionTime &
\native{75.21 $\pm$ 0.86} &
\best{\calib{82.67 $\pm$ 0.49}} &
\second{\calib{82.19 $\pm$ 1.20}} &
\calib{82.01 $\pm$ 0.59} &
+7.46 \\
\bottomrule
\end{tabular}
}
\end{subtable}

\vspace{1.2em}

% ============================================================
% h = 20
% ============================================================
\begin{subtable}{\textwidth}
\centering
\caption{$h=20$}
\resizebox{\textwidth}{!}{
\begin{tabular}{lccccc}
\toprule
Model & Native & MS & ZOnly & RCG & $\Delta_{\mathrm{best}}$ \\
\midrule
BiNCTABL &
\native{73.63 $\pm$ 0.28} &
\best{\calib{82.61 $\pm$ 0.39}} &
\calib{81.69 $\pm$ 0.26} &
\second{\calib{81.81 $\pm$ 0.37}} &
+8.99 \\
TLOB &
\native{69.53 $\pm$ 0.29} &
\calib{79.31 $\pm$ 1.07} &
\second{\calib{79.51 $\pm$ 0.45}} &
\best{\calib{80.17 $\pm$ 0.47}} &
+10.65 \\
InceptionTime &
\native{71.85 $\pm$ 1.91} &
\best{\calib{80.74 $\pm$ 0.74}} &
\second{\calib{80.37 $\pm$ 0.78}} &
\calib{79.86 $\pm$ 0.93} &
+8.89 \\
\bottomrule
\end{tabular}
}
\end{subtable}

\vspace{1.2em}

% ============================================================
% h = 50
% ============================================================
\begin{subtable}{\textwidth}
\centering
\caption{$h=50$}
\resizebox{\textwidth}{!}{
\begin{tabular}{lccccc}
\toprule
Model & Native & MS & ZOnly & RCG & $\Delta_{\mathrm{best}}$ \\
\midrule
BiNCTABL &
\native{88.64 $\pm$ 0.20} &
\best{\calib{92.12 $\pm$ 0.17}} &
\calib{91.93 $\pm$ 0.25} &
\second{\calib{92.07 $\pm$ 0.22}} &
+3.48 \\
TLOB &
\native{81.78 $\pm$ 0.39} &
\best{\calib{88.37 $\pm$ 0.17}} &
\second{\calib{88.04 $\pm$ 0.32}} &
\calib{87.86 $\pm$ 0.28} &
+6.59 \\
InceptionTime &
\native{82.26 $\pm$ 1.56} &
\best{\calib{88.21 $\pm$ 0.49}} &
\second{\calib{87.85 $\pm$ 0.45}} &
\calib{87.55 $\pm$ 0.56} &
+5.95 \\
\bottomrule
\end{tabular}
}
\end{subtable}

\vspace{1.2em}

% ============================================================
% h = 100
% ============================================================
\begin{subtable}{\textwidth}
\centering
\caption{$h=100$}
\resizebox{\textwidth}{!}{
\begin{tabular}{lccccc}
\toprule
Model & Native & MS & ZOnly & RCG & $\Delta_{\mathrm{best}}$ \\
\midrule
BiNCTABL &
\native{93.18 $\pm$ 0.11} &
\calib{94.95 $\pm$ 0.22} &
\second{\calib{94.95 $\pm$ 0.11}} &
\best{\calib{95.01 $\pm$ 0.26}} &
+1.83 \\
TLOB &
\native{85.44 $\pm$ 0.26} &
\second{\calib{90.22 $\pm$ 0.22}} &
\calib{90.09 $\pm$ 0.14} &
\best{\calib{90.44 $\pm$ 0.24}} &
+5.00 \\
InceptionTime &
\native{85.22 $\pm$ 1.03} &
\best{\calib{89.99 $\pm$ 0.23}} &
\second{\calib{89.48 $\pm$ 0.37}} &
\calib{89.37 $\pm$ 0.37} &
+4.77 \\
\bottomrule
\end{tabular}
}
\end{subtable}

\vspace{0.6em}
\normalsize

\end{table*}

\section{Impact and Applications}
\label{app:possible_applications}

The proposed framework is not tied to a single temporal domain. Its main use is to diagnose whether the remaining errors of a trained temporal classifier come from missing temporal evidence, underused decision-level evidence, or limited remaining headroom. This makes the method relevant whenever a deployed or pretrained temporal model is already available and retraining the full backbone is expensive, risky, or undesirable.

\paragraph{High-frequency financial prediction.}
Limit-order-book forecasting is a natural application because market microstructure contains information at multiple temporal resolutions. Very short windows may capture local order-flow imbalance, while coarser views may capture more stable liquidity and trend context. The FI-2010 results suggest that residual multi-scale evidence is especially useful in short-horizon, high-noise settings, where a native decision boundary may not fully resolve rapidly changing temporal patterns. In such settings, a conservative residual branch can improve the predictor without replacing the specialized backbone. However, financial deployment would require additional validation under transaction costs, market impact, latency constraints, regime shifts, and strict risk controls. The method should therefore be viewed as a forecasting and diagnostic component, not as a complete trading system.

\paragraph{Clinical temporal classification.}
Clinical time series, including ECG and physiological monitoring, often contain evidence at multiple scales: short local events, medium-range rhythm patterns, and broader contextual changes. A representation--calibration decomposition can be useful when a strong clinical classifier already exists but its decisions remain sensitive to thresholds or class imbalance. The residual branch can test whether additional temporal evidence is missing, while post-hoc calibration can test whether the available logits are being converted into decisions in a way that matches the target clinical metric. In this setting, the method could be used as an audit tool for model improvement. It should not be interpreted as sufficient for clinical deployment without prospective validation, subgroup analysis, uncertainty assessment, and human expert review.

\paragraph{Wearable sensing and activity recognition.}
Wearable activity-recognition systems often operate under sensor noise, subject variability, device placement changes, and heterogeneous movement patterns. These conditions make trust allocation across temporal scales relevant: fine-scale signals may capture rapid movements, while coarser views may be more robust to local noise. The results on HARTH, MHEALTH, and UCI-HAR suggest that the method is most useful when the native model is not already saturated, especially for weaker recurrent backbones or unstable datasets. In practical wearable systems, the framework could help identify whether performance limitations come from insufficient temporal representation or from poorly calibrated decision boundaries.

\paragraph{Industrial monitoring and predictive maintenance.}
Many industrial systems produce multivariate temporal streams from sensors, machines, or infrastructure. Fault signatures may appear as short-lived spikes, persistent drifts, or multi-scale patterns. A conservative residual adapter can be attached to an existing classifier to test whether multi-scale evidence improves fault detection without retraining the entire monitoring system. Post-hoc calibration can then adjust the decision rule to better match operational costs, such as false alarms versus missed failures. This is particularly relevant when models must be updated cautiously because of safety, certification, or maintenance constraints.

\paragraph{Human-in-the-loop model auditing.}
Beyond direct performance improvement, the framework can be used as a diagnostic tool. If residual branches improve the raw predictor, the native model likely lacks useful temporal evidence. If raw residual gains are small but branch-aware calibration improves performance, the evidence may already be present but underused at the decision layer. If neither helps, the model may be close to saturation for the available data and backbone. This diagnostic interpretation can guide practitioners toward different interventions: collecting more data, improving temporal encoders, adjusting decision thresholds, or focusing on calibration rather than representation learning.

\paragraph{Responsible use.}
The same properties that make the framework useful also require caution. In clinical or wearable applications, improved temporal classification can affect decisions about health, activity, or behavior, and errors may have uneven consequences across groups. In financial applications, improvements in high-frequency prediction may interact with market dynamics and should not be evaluated only by classification metrics. More generally, calibration can improve a target metric without guaranteeing fairness, causal validity, robustness under distribution shift, or reliable uncertainty estimates. For this reason, any deployment should include domain-specific validation, monitoring under distribution shift, subgroup-level evaluation when relevant, and clear separation between diagnostic use and automated decision-making.

\section{Limitations and Future Work}
\label{app:limitations}

This study has several limitations. First, the empirical evaluation is centered on a restricted set of temporal classification benchmarks and backbone families. Although the selected datasets cover high-frequency financial prediction, ECG classification, and wearable activity recognition, broader validation across additional domains, architectures, sequence lengths, and class-imbalance regimes would strengthen the generality of the conclusions. Second, all experiments use three seeds. This is common in deep temporal learning studies, but it still limits the precision of variance estimates, especially on unstable datasets such as MHEALTH and on settings where calibration can be sensitive to validation-set fluctuations.

Third, the proposed calibration stage is post-hoc: it changes only the decision rule after representation learning is frozen. This design is useful for isolating decision-level effects, but it does not answer whether tighter end-to-end integration between residual trust allocation and calibration could yield stronger results. Fourth, the residual branch is intentionally conservative, but its design space is only partially explored. Other reliability descriptors, scale-fusion mechanisms, auxiliary encoders, or regularization strategies may produce more robust variants, particularly in regimes where explicit reliability information is useful.

Finally, the decomposition explains where gains appear to come from---native representation, residual correction, or decision calibration---but it does not yet provide a full theoretical account of when each component should dominate. Developing predictive criteria for distinguishing representation-opportunity, calibration-opportunity, and near-saturation regimes is an important direction for future work. More broadly, the results suggest that progress in temporal classification should not be framed only as building stronger feature extractors, but also as understanding how evidence from multiple temporal views should be trusted and combined at decision time.

\section{Additional Related Work}
\label{app:additional_related_work}

\paragraph{Additional temporal encoders.}
The temporal modeling literature includes recurrent, convolutional, self-supervised, and Transformer-based approaches. LSTMs introduced gated recurrence as a mechanism for learning long-range dependencies in sequential data~\citep{lstm}, while temporal convolutional networks showed that convolutional sequence models can be competitive alternatives to recurrent architectures for generic sequence modeling~\citep{bai2018empiricalevaluationgenericconvolutional}. Fully convolutional networks also became a strong baseline for end-to-end time-series classification~\citep{fcn}. Beyond supervised learning, TS2Vec introduced a general contrastive framework for time-series representation learning~\citep{yue2022ts2vec}. Recent architectures further explore interaction-based downsampling, multi-periodic temporal structure, and inverted tokenization for time-series modeling~\citep{NEURIPS2022_266983d0,wu2022timesnet,liu2023itransformer}. These works motivate the use of strong native temporal predictors, but they do not directly isolate whether residual errors are representational or decisional.

\paragraph{Adapters, ensembles, and auxiliary prediction heads.}
Our residual branch is related to adapter-style methods and output-level combination, but differs in its diagnostic use. Residual adapters add small trainable modules to an existing network while preserving much of the original model~\citep{NIPS2017_e7b24b11}, and parameter-efficient adapters similarly adapt large pretrained models while keeping the backbone fixed~\citep{pmlr-v97-houlsby19a}. Deep ensembles combine multiple predictors to improve robustness and uncertainty estimation~\citep{lakshminarayanan2017simple}. In contrast, our auxiliary branch is not intended to form a conventional ensemble. It is constrained to provide residual temporal evidence, and the final calibrator receives branch-separated native and residual logits. This preserves the distinction between evidence construction and decision conversion.

\paragraph{Classical calibration and test-time adaptation.}
Classical calibration methods transform classifier scores into better probability estimates. Zadrozny and Elkan proposed multiclass calibration from classifier scores~\citep{10.1145/775047.775151}, while Niculescu-Mizil and Caruana analyzed how different supervised learning algorithms distort probability estimates and how post-hoc calibration can correct them~\citep{10.1145/1102351.1102430}. Test-time adaptation methods pursue a different goal: they modify the model at deployment to improve robustness under shift. MEMO, for example, adapts predictions through augmentation-based test-time optimization~\citep{zhang2022memo}. Our calibration stage is more restrictive: it does not use unlabeled test examples and does not update the backbone. This restriction is intentional, because it makes the post-hoc stage interpretable as decision-level adjustment over already-learned temporal evidence.

\paragraph{Benchmark domains.}
Our experiments cover high-frequency financial prediction, ECG classification, and wearable human-activity recognition. FI-2010 is a standard benchmark for mid-price forecasting from limit-order-book data~\citep{fi2010}. Specialized models for this domain include DeepLOB~\citep{zhang2019deeplob}, normalization and bilinear structures for high-frequency financial time series~\citep{tran2021data}, and TLOB~\citep{tlob}. PTB-XL is a large public 12-lead ECG dataset for clinical ECG classification~\citep{ptbxl}. UCI-HAR, MHEALTH, and HARTH provide wearable-sensor benchmarks for human-activity recognition under different sensing and annotation conditions~\citep{human_activity_recognition_using_smartphones_240,mhealth_319,harth}. These datasets are useful for our regime analysis because they differ substantially in temporal resolution, noise, class structure, and saturation level.

%%%%%%%%%%%%%%%%%%%%%%%%%%%%%%%%%%%%%%%%%%%%%%%%%%%%%%%%%%%%%%

\newpage
\section*{NeurIPS Paper Checklist}

\begin{enumerate}

\item {\bf Claims}
    \item[] Question: Do the main claims made in the abstract and introduction accurately reflect the paper's contributions and scope?
    \item[] Answer: \answerYes{}.
    \item[] Justification: The abstract and introduction state the paper's main scope: a representation--calibration decomposition for temporal classification, implemented through frozen native predictors, residual multi-scale corrections, and post-hoc decision calibration. The claims are presented as regime-dependent rather than universal, and the experimental section and limitations discuss when residual evidence or calibration helps, when gains are marginal, and where the conclusions may not generalize.

\item {\bf Limitations}
    \item[] Question: Does the paper discuss the limitations of the work performed by the authors?
    \item[] Answer: \answerYes{}.
    \item[] Justification: The paper discusses limitations in Section~\ref{sec:conclusion}, including the restricted set of datasets and backbones, the use of three seeds, the post-hoc nature of the calibration layer, the partially explored design space of residual adapters and reliability descriptors, and the lack of a theoretical account predicting when each regime should dominate. Additional implementation and compute-related details are provided in Appendix~\ref{app:experimental_setup}.

\item {\bf Theory assumptions and proofs}
    \item[] Question: For each theoretical result, does the paper provide the full set of assumptions and a complete (and correct) proof?
    \item[] Answer: \answerNA{}.
    \item[] Justification: The paper does not present formal theoretical results, theorems, lemmas, or proofs. The mathematical content is used to define the model decomposition, residual adapter, training objectives, and post-hoc calibration procedures, while the main claims are evaluated empirically.

\item {\bf Experimental result reproducibility}
    \item[] Question: Does the paper fully disclose all the information needed to reproduce the main experimental results of the paper to the extent that it affects the main claims and/or conclusions of the paper (regardless of whether the code and data are provided or not)?
    \item[] Answer: \answerYes{}.
    \item[] Justification: The paper describes the proposed decomposition, residual adapters, calibration methods, datasets, backbone families, evaluation metrics, seeds, and main experimental protocol in Sections~\ref{sec:experiments} and~\ref{app:experimental_setup}. The appendix further specifies training hyperparameters, scale factors, optimizer settings, calibration search spaces, bootstrap evaluation, and saved artifacts, while the supplementary material provides code and running instructions for reproducing the reported results.

\item {\bf Open access to data and code}
    \item[] Question: Does the paper provide open access to the data and code, with sufficient instructions to faithfully reproduce the main experimental results, as described in supplemental material?
    \item[] Answer: \answerYes{}.
    \item[] Justification: The paper uses publicly available benchmark datasets and provides code, data-access instructions, and running instructions in the Supplementary Material. Appendix~\ref{app:experimental_setup} describes the experimental setup required to reproduce the main results, including data preparation assumptions, model variants, training hyperparameters, calibration procedures, evaluation metrics, random seeds, and saved output artifacts.

\item {\bf Experimental setting/details}
    \item[] Question: Does the paper specify all the training and test details (e.g., data splits, hyperparameters, how they were chosen, type of optimizer) necessary to understand the results?
    \item[] Answer: \answerYes{}.
    \item[] Justification: The experimental protocol is described in Section~\ref{sec:experiments}, including the datasets, backbone families, residual variants, calibration methods, primary and secondary metrics, and the use of three random seeds. Appendix~\ref{app:experimental_setup} provides the remaining training and evaluation details, including optimizer, learning rate, weight decay, batch size, number of epochs, early stopping, scale factors, calibration search spaces, bootstrap protocol, and saved artifacts; dataset access, split construction, and running commands are provided in the Supplementary Material.

\item {\bf Experiment statistical significance}
    \item[] Question: Does the paper report error bars suitably and correctly defined or other appropriate information about the statistical significance of the experiments?
    \item[] Answer: \answerYes{}.
    \item[] Justification: The main results report mean $\pm$ standard deviation over three random seeds, making the source of variability explicit in Table~\ref{tab:dataset_residual_macro_f1}. Appendix~\ref{app:experimental_setup} further describes the paired bootstrap protocol used for model comparisons, including the number of bootstrap samples, confidence interval quantiles, and the empirical probability \(p(\Delta>0)\).

\item {\bf Experiments compute resources}
    \item[] Question: For each experiment, does the paper provide sufficient information on the computer resources (type of compute workers, memory, time of execution) needed to reproduce the experiments?
    \item[] Answer: \answerYes{} % Replace by \answerYes{}, \answerNo{}, or \answerNA{}.
    \item[] Justification: The paper reports the computational environment used for the experimental runs, including the operating system, CPU, system memory, GPU model and GPU memory, CUDA/NVIDIA driver versions, Python version, and local NVMe-backed storage. The appendix further specifies the compute-relevant training setup for each experimental family, including random seeds, batch sizes, optimizer, learning rate, weight decay, gradient clipping, dropout, early-stopping patience, maximum numbers of epochs for native/baseline models and residual adapters, residual model dimensions, scale factors, and calibration search budgets.
    
\item {\bf Code of ethics}
    \item[] Question: Does the research conducted in the paper conform, in every respect, with the NeurIPS Code of Ethics \url{https://neurips.cc/public/EthicsGuidelines}?
    \item[] Answer: \answerYes{}.
    \item[] Justification: The authors have reviewed the NeurIPS Code of Ethics and, to the best of their knowledge, the research conforms to it. The work uses established benchmark datasets, does not involve human-subject experiments, private data collection, scraped datasets, or deployment affecting individuals, and the submission is prepared to preserve anonymity during review.

\item {\bf Broader impacts}
    \item[] Question: Does the paper discuss both potential positive societal impacts and negative societal impacts of the work performed?
    \item[] Answer: \answerYes{} % Replace by \answerYes{}, \answerNo{}, or \answerNA{}.
    \item[] Justification: The paper discusses a foundational inference-time method for temporal classification, but the application domains evaluated in the work include ECG classification, wearable human-activity recognition, and limit-order-book forecasting. These domains create plausible societal impacts even though the paper does not deploy a real-world system. The possible impacts of the paper are discussed in Appendix~\ref{app:possible_applications}.
    
\item {\bf Safeguards}
    \item[] Question: Does the paper describe safeguards that have been put in place for responsible release of data or models that have a high risk for misuse (e.g., pre-trained language models, image generators, or scraped datasets)?
    \item[] Answer: \answerNA{}.
    \item[] Justification: The paper does not release high-risk models, generative systems, scraped datasets, or assets with substantial dual-use concerns. The released materials are limited to code, experimental configurations, and results for temporal classification on established benchmark datasets, so no additional safeguards of the type described in the question are required.

\item {\bf Licenses for existing assets}
    \item[] Question: Are the creators or original owners of assets (e.g., code, data, models), used in the paper, properly credited and are the license and terms of use explicitly mentioned and properly respected?
    \item[] Answer: \answerYes{} % Replace by \answerYes{}, \answerNo{}, or \answerNA{}.
    \item[] Justification: The paper uses existing public benchmark datasets and standard model architectures. The original creators are credited through citations, and the supplementary material provides an asset table listing each dataset/code asset, its source, version or access date where applicable, license or terms of use, and any required attribution. The authors use these assets only for benchmark evaluation and respect their stated licenses and usage conditions.

\item {\bf New assets}
    \item[] Question: Are new assets introduced in the paper well documented and is the documentation provided alongside the assets?
    \item[] Answer: \answerYes{}.
    \item[] Justification: The paper does not introduce a new dataset, but it does release new code, experimental configurations, and generated experimental artifacts for reproducing the proposed residual-calibration framework. These assets are documented in the Supplementary Material, while Appendix~\ref{app:experimental_setup} describes the training setup, calibration procedures, saved outputs, hyperparameters, evaluation metrics, and reproducibility details.

\item {\bf Crowdsourcing and research with human subjects}
    \item[] Question: For crowdsourcing experiments and research with human subjects, does the paper include the full text of instructions given to participants and screenshots, if applicable, as well as details about compensation (if any)? 
    \item[] Answer: \answerNA{}.
    \item[] Justification: The paper does not involve crowdsourcing, user studies, interviews, experiments with human participants, participant instructions, screenshots, or compensation. The empirical evaluation is conducted only on established temporal classification benchmark datasets, so this item is not applicable.

\item {\bf Institutional review board (IRB) approvals or equivalent for research with human subjects}
    \item[] Question: Does the paper describe potential risks incurred by study participants, whether such risks were disclosed to the subjects, and whether Institutional Review Board (IRB) approvals (or an equivalent approval/review based on the requirements of your country or institution) were obtained?
    \item[] Answer: \answerNA{}.
    \item[] Justification: The paper does not involve crowdsourcing, user studies, interviews, experiments with human participants, or the collection of new human-subject data. The empirical evaluation is conducted only on established temporal classification benchmark datasets, so IRB approval or equivalent human-subjects review is not applicable.

\item {\bf Declaration of LLM usage}
    \item[] Question: Does the paper describe the usage of LLMs if it is an important, original, or non-standard component of the core methods in this research? Note that if the LLM is used only for writing, editing, or formatting purposes and does \emph{not} impact the core methodology, scientific rigor, or originality of the research, declaration is not required.
    \item[] Answer: \answerNA{}.
    \item[] Justification: LLMs were not used as an important, original, or non-standard component of the core research methodology, experiments, model design, data processing, or analysis. They were used only for writing support, including improving clarity, grammar, wording, and formatting; all scientific ideas, experimental design, implementation, results, and conclusions were produced and verified by the authors.

\end{enumerate}

\end{document}